%% file: paper.tex
\documentclass{article}

\PassOptionsToPackage{numbers, compress}{natbib}

    \usepackage[preprint]{neurips_2023}

\usepackage[utf8]{inputenc} 
\usepackage[T1]{fontenc}    
\usepackage{hyperref}       
\usepackage{url}            
\usepackage{booktabs}       
\usepackage{amsfonts}       
\usepackage{nicefrac}       
\usepackage{microtype}      
\usepackage{xcolor}         
\usepackage{graphicx}
\usepackage{amsmath}
\usepackage{dsfont}
\usepackage{cleveref}
\usepackage{soul}
\usepackage{hyphenat}

\definecolor{mediumblue}{rgb}{0.2,0.6,0.9}
\definecolor{lightblue}{rgb}{0.8,0.9,1}
\sethlcolor{lightblue}

\newcommand\kl[1]{D_\mathrm{KL}(#1)}
\newcommand\newlinesymbol{\textcolor{lightgray}{\textbackslash n}}

\title{Likelihood-Based Diffusion Language Models}

\author{%
  Ishaan Gulrajani \\
  Stanford University\\
  \texttt{igul222@gmail.com} \\
  \And
  Tatsunori B. Hashimoto \\
  Stanford University\\
  \texttt{thashim@stanford.edu} \\
}

\begin{document}

\maketitle

\begin{abstract}
Despite a growing interest in diffusion-based language models, existing work has not shown that these models can attain nontrivial likelihoods on standard language modeling benchmarks.
In this work, we take the first steps towards closing the likelihood gap between autoregressive and diffusion-based language models, with the goal of building and releasing a diffusion model which outperforms a small but widely-known autoregressive model.
We pursue this goal through algorithmic improvements, scaling laws, and increased compute.
On the algorithmic front, we introduce several methodological improvements for the maximum-likelihood training of diffusion language models.
We then study scaling laws for our diffusion models and find compute-optimal training regimes which differ substantially from autoregressive models.
Using our methods and scaling analysis, we train and release Plaid 1B, a large diffusion language model which outperforms GPT-2 124M in likelihood on benchmark datasets and generates fluent samples in unconditional and zero-shot control settings.
\footnote{We release our code and pretrained models at \texttt{https://github.com/igul222/plaid}.}
\end{abstract}

\input{sections/intro}
\input{sections/background}
\input{sections/method}
\input{sections/ablations}
\input{sections/scaling}
\input{sections/plaid1b}
\input{sections/related}
\input{sections/conclusion}

\newpage
\bibliographystyle{plainnat}
\bibliography{tex-common/all.bib,paper.bib}

\newpage
\appendix


\input{sections/appendix_weight_schedules}
\input{sections/appendix_experiment_details}
\input{sections/appendix_isoflop_profiles}
\input{sections/appendix_samples}

\end{document}

%% file: sections/intro.tex
\section{Introduction}

Large language models lie at the center of recent advances in artificial intelligence.
Shared across nearly all such language models is a common recipe: learn a model that maximizes data likelihoods using an autoregressive, left-to-right factorization.
Maximum-likelihood pretraining has been a remarkably successful paradigm, leading to models that perform well on a range of downstream tasks and display complex behaviors like in-context learning~\cite{Brown20,openai2023gpt4}.

Thus far, autoregressive modeling has been a core part of this process due to its computational efficiency and empirical performance.
However, this choice carries drawbacks.
Autoregressive models generate tokens one at a time, making it difficult to perform long-range planning or controllable generation \cite{Keskar19,Krause20,Li22}.
In addition, certain sequence distributions may be fundamentally more difficult to model autoregressively \cite{Lin20}.

Given the importance of language modeling, these potential drawbacks motivate us to explore alternatives to the autoregressive approach.
As a promising candidate, we turn to continuous diffusion models \cite{SohlDickst15,Ho20}, which have achieved state-of-the-art results in image modeling \cite{Dhariwal21,Rombach21,Saharia22}.
In language, prior works on diffusion models exist \cite[e.g.][]{Li22,han2022ssd,Dieleman22}, but these optimize non-likelihood-based objectives.
Without the ability to use standard likelihood-based benchmarks~\cite{Marcus99,hutterprize,Merity16}, it is difficult to say precisely how these models compare to autoregressive models (see \Cref{sec:related_work} for a discussion).
Somewhat concerningly, there is no work showing that it is possible for diffusion language models to achieve any nontrivial likelihoods on standard benchmarks.

In this work, we explore the limits of likelihood-based diffusion language models. 
Our goal is to train and release a diffusion model which achieves better likelihoods than GPT-2 124M \cite{Radford19}, which we consider the smallest widely-adopted autoregressive model today.
To achieve this goal, we first develop an algorithmic framework, then study its scaling laws to enable compute-optimal training, and finally train a large model called Plaid 1B.

Our contributions are as follows:
\begin{enumerate}
    \item
      We explore the design space of likelihood-based diffusion language models and propose an algorithmic framework called Plaid.
      We validate the design choices of Plaid through compute-matched ablations.
    \item 
      We study the scaling laws of Plaid training.
      Our analysis shows that the log-likelihood of Plaid models improves predictably with more compute.
      We derive a recipe for compute-optimal training which differs substantially from the usual autoregressive rule.
    \item
      We train and release Plaid 1B, a large diffusion language model pretrained on OpenWebText2~\cite{pile}.
      Plaid 1B outperforms GPT-2 124M in zero-shot likelihood across six standard benchmarks.
      We demonstrate Plaid 1B's ability to perform fluent and controllable text generation.      
  \end{enumerate}

\begin{figure}[t!]
\centering
\includegraphics[width=4in]{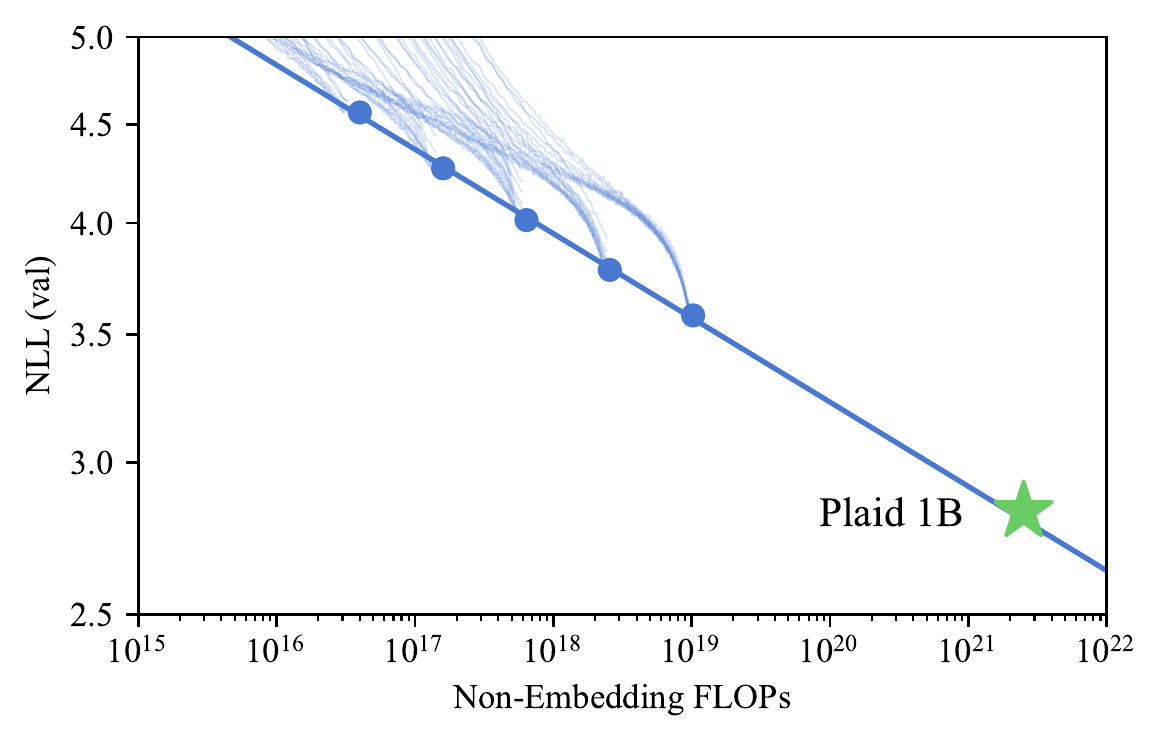}
\caption{
Plaid models scale predictably across five orders of magnitude.
Our largest model, Plaid 1B, outperforms GPT-2 124M in zero-shot likelihood (see \Cref{tab:likelihood}).}
\label{fig:figure1}
\end{figure}


%% file: sections/background.tex
\section{Variational Diffusion Models for language}
\label{sec:background}

In this background section, we formally define continuous diffusion models over text sequences, adopting the Variational Diffusion Models (VDM) framework \cite{Kingma21} which is a natural fit for likelihood-based training (see \citet{Karras22} for a survey on other formalisms).
For brevity, we simplify some details in our exposition and refer the reader to \citet{Kingma21} for details.

Consistent with prior work (e.g. \citet{Li22}), our basic approach will be to map discrete text sequences into a continuous space with a token-wise embedding function and then construct a diffusion model on the embedded data.

\subsection{Forward diffusion process}

Consider a sequence of tokens $x=[x^{(1)}, \ldots, x^{(L)}]$ drawn from the data distribution $q(x)$.
We transform $x$ into a sequence $\tilde{x}$ of embedding vectors using an invertible token-wise embedding function $\mathrm{Embed}(\cdot)$, such that $\tilde{x}^{(i)} := \mathrm{Embed}(x^{(i)})$.

The \textit{forward process} is a Markov chain over latent variables $z_t$ from $t=0$ to $t=1$ which progressively adds Gaussian noise to $\tilde{x}$.
Let $\sigma^2(t)$ be some monotonic function that specifies the total noise added by time $t$.
We then define the forward process distribution $q$ with $T$ discrete timesteps as
\begin{align}
q(x,z) := q(x) q(z_0 | x) \prod_{i=1}^{T} q(z_{i/T}|z_{(i-1)/T})
\end{align}
where $q(z_0 | x):=\mathcal{N}(\tilde{x}, \sigma^2(0))$ and $q(z_t|z_s):=\mathcal{N}(z_s, \sigma^2(t) - \sigma^2(s))$.
It follows from this that $q(z_s | z_t, \tilde{x})$ is also Gaussian, which will be useful later.

\subsection{Reverse generative process}

We can approximate the forward process distribution $q$ by a learned Markov \textit{reverse process} where time runs backward from $t=1$ to $t=0$.
The reverse process with $T$ timesteps is defined via the decomposition
\begin{align}
p_\theta(x, z) := p(z_1) \left( \prod_{i=1}^{T} p_\theta(z_{(i-1)/T} | z_{i/T}) \right) p(x | z_0).
\end{align}
Let $z_t^{(i)}$ denote the portion of $z_t$ at sequence position $i$.
Then we choose $p(z_1) := \mathcal{N}(0, \sigma^2(1)I)$ and $p(x | z_0) := \prod_i p(x^{(i)} | z^{(i)}_{0})$ where $p(x^{(i)} | z^{(i)}_{0}) \propto q(z^{(i)}_{0}|x_i)$.
Recalling that $q(z_{(i-1)/T} | z_{i/T}, \tilde{x})$ is Gaussian, for the remaining factors we choose $p_\theta(z_{(i-1)/T} | z_{i/T}) := q(z_{(i-1)/T} | z_{i/T}, \tilde{x}=\hat{x}_\theta(z_{i/T}))$ where $\hat{x}_\theta(z_t)$ is a \textit{denoiser} neural network that approximates $\mathbb{E}_q[\tilde{x}|z_t]$.
Finally, our generative model is given by the marginal distribution $p_\theta(x) = \int_z p_\theta(x,z)$.
If $\hat{x}_\theta$ is optimal, then the forward and reverse processes express the same joint distribution as $\sigma^2(0) \rightarrow 0$, $\sigma^2(1) \rightarrow \infty$, and $T \rightarrow \infty$.

\subsection{Likelihood bound}

To optimize and evaluate the likelihood, we can write a variational lower bound (VLB) for the log-likelihood as
\begin{align}
-\log p_\theta(x) \leq -\mathrm{VLB}(x) := \kl{q(z_1|x) \| p(z_1)} \;+\; \mathbb{E}_{q(z_0|x)}[-\log p(x|z_0)] \;+\; \mathcal{L}_T \label{eq:loss}
\end{align}
where
\begin{align}
\mathcal{L}_T := \sum_{i=1}^{T} \mathbb{E}_{q(z_{i/T}|x)}[ \kl{q(z_{(i-1)/T} | z_{i/T}, x) \| p_\theta(z_{(i-1)/T} | z_{i/T})}].
\end{align}
In the $T \rightarrow \infty$ limit, $\mathcal{L}_T$ simplifies to
\begin{align}
\mathcal{L}_\infty = - \frac{1}{2} \mathbb{E}_{t \sim U[0,1], z_t \sim q(z_t|x)}[ \mathrm{SNR}'(t) \| x - \hat{x}_\theta(z_t) \|_2^2 ]
\end{align}
where $\mathrm{SNR}'(t) := \frac{d}{dt} \frac{1}{\sigma^2(t)}$.
We use Monte-Carlo estimates of the resulting continuous-time likelihood bound to train and evaluate our model.

\subsection{Learned noise schedule}

A crucial hyperparameter in diffusion models is the noise schedule $\sigma^2(t)$, which specifies how much noise to add at each time in the diffusion process.
In our setting, the VLB is differentiable with respect to $\sigma^2(t)$ via the reparameterization trick.
Moreover, the VLB is invariant to the value of $\sigma^2(t)$ except at $t=0$ and $t=1$ in the continuous-time limit.

We can therefore parameterize $\sigma^2(t)$ as a scalar-to-scalar neural network and learn it by gradient descent.
We train the endpoints $\sigma^2(0)$ and $\sigma^2(1)$ to maximize the VLB, and the schedule in between the endpoints to minimize the variance of the Monte-Carlo estimate of the VLB.
Minimizing the loss variance is a proxy for minimizing the gradient covariance trace, which generally speeds up learning.
See \citet{Kingma21} for further implementation details about this training procedure.


%% file: sections/method.tex
\section{The Plaid framework}
\label{sec:method}

In this section, we present a series of algorithmic improvements to the basic setup described in \Cref{sec:background}.
The result is a framework for diffusion language models which we refer to as Plaid (Perplexity-based LAnguage Inverse Diffusion).

\subsection{Learned embeddings}

In an autoregressive language model, the embedding operation is simply the first layer of the neural network and thus can be treated as just another part of the network.
This is not true of embeddings in diffusion language models, which play a more fundamental role: they determine the order in which different tokens get generated.
Tokens whose embeddings are far apart become distinguishable early in the reverse process, whereas nearby embeddings are distinguishable only later, at low noise levels.

Despite the importance of embeddings in diffusion language models, the loss functions used in prior work \cite{Li22,Dieleman22} lead to ill-posed problems when optimized over $W_{\text{Embed}}$: for example, if our objective is $L_2$ reconstruction, then collapsing the embeddings by setting $W_{\text{Embed}} = 0$ and $\hat{x}_\theta(z_t) = 0$ yields a degenerate solution with zero loss.
Prior work addresses this with workarounds like choosing $W_{\mathrm{Embed}}$ by hand \cite{Chen22,strudel2022self} or using heuristic regularizers \cite{Li22} or constraints \cite{Dieleman22}.

In contrast, the Plaid loss function is a bound on the log-likelihood of the discrete data, which is a meaningful objective over both the model weights and embeddings.
We therefore optimize the embedding matrix $W_{\mathrm{Embed}}$ jointly with the rest of the model without additional constraints.

\subsection{Categorical reparameterization}
\label{sec:categorical_reparameterization}

When optimally trained, $\hat{x}_\theta(z_t)$ learns to approximate a conditional expectation $\mathbb{E}[\tilde{x}|z_t]$ over sequences of word embeddings $\tilde{x}$.
At low noise levels, some or all of the embeddings in $\tilde{x}$ are deterministic given $z_t$, so an optimal $\hat{x}_\theta(z_t)$ should output these exactly.
However, doing so requires memorizing embedding vectors to high precision somewhere inside the model parameters, which is a poor use of capacity.

Instead of forcing the model to memorize the embedding vectors, we reparameterize $\hat{x}_\theta(z_t)$ as an average of embeddings weighted by a softmax over tokens.
More formally, let $f_\theta(z_t)$ be a neural network which outputs logits and define $\hat{x}$ as an average over embeddings $\hat{x}_\theta^{(i)}(z_t) := W_{\mathrm{Embed}}\mathrm{softmax}(f_\theta^{(i)}(z_t))$.
We can interpret $f$ as learning a posterior over each discrete token $x^{(i)}$ given $z_t$.
This relates to methods proposed in prior work, but these either require proxy objectives \cite{Li22,Dieleman22} or consider image models \cite{Chen22}.

\subsection{Output prior}

When we interpret $f_\theta$ as a posterior over tokens, the optimal value of $f_\theta(z_t)$ is $\log q(x^{(i)}|z_t) + Z$, which decomposes as $\log q(z_t^{(i)}|x^{(i)}) + \log q(x^{(i)} | z^{(\neq i)}_t) + Z$ where $z^{(\neq i)}_t := \{z^{(j)}_t : j \neq i \}$.
We view the first term as a prior constraining the model's predictions to those which are plausible given $z_t$, while the second term models relationships between different tokens.

To make it easier to model $f_\theta$, we compute the first term in closed form as the log-density of a Gaussian $\mathcal{N}(\tilde{x}^{(i)}, \sigma^2(t)I)$ and add it to the output.
This leaves the neural network with only the easier task of estimating $\log p(x^{(i)} | z^{(\neq i)}_t)$.
Empirically, we found it helpful to linearly anneal in this prior over the first 5000 steps of training.

\subsection{Learned conditional likelihood}

Recall that our loss function \eqref{eq:loss} includes a conditional likelihood term $\log p(x|z_0)$.
We are free to choose $p$ however we wish, and in~\Cref{sec:background} we chose a position-wise factorial model $p(x | z_0) := \prod_i p(x^{(i)} | z^{(i)}_{0})$, with a simple fixed distribution for each factor.
This choice is optimal for sufficiently small $\sigma^2(0)$, but using a more powerful model allows $\sigma^2(0)$ to take a larger value, effectively truncating the reverse process and therefore making it simpler to learn.

Here we leverage the fact that, after applying the categorical reparameterization (\Cref{sec:categorical_reparameterization}), our neural network $f_\theta(z_t)$ can be interpreted as learning the logits for $q(x^{(i)}|z_t)$ at all positions $i$.
We therefore choose to keep $p(x|z_0)$ as a factorial model, but define each factor $p(x_i | z^{(i)}_{0})$ using the more powerful learned model $\mathrm{softmax}(f^{(i)}_\theta(z_t))$.

Implementing this change naively requires two evaluations of $f_\theta$ for each minibatch example during training, corresponding to the two terms of \eqref{eq:loss} $\mathcal{L}_\infty$ and $\log p_\theta(x|z_0)$.
We instead split each minibatch, using some examples to compute $\mathcal{L}_\infty$ and the rest to compute $\log p_\theta(x|z_0)$.
We allocate examples between the two terms according to the ratio $\sqrt\frac{\mathrm{Var}(\mathcal{L}_\infty)}{\mathrm{Var}(\log p_\theta(x|z_0))}$, where we compute the variances using running estimates of each term's first and second moments.
This minimizes the variance of the full loss \eqref{eq:loss}.

\subsection{Self-conditioning}

Self-conditioning \cite{Chen22} is a technique which improve the performance of diffusion language models.
The core idea is to reparameterize the denoiser $\hat{x}_\theta(z_t)$ as the fixed point $y_\infty$ of a recurrence $y_0 := 0, y_{i+1} := \hat{x}_\theta'(z_t, y_i)$ where $\hat{x}_\theta'$ is a neural network which now takes two inputs instead of one.
During training, we approximate the fixed point $y_\infty$ by randomly unrolling the recurrence to either $y_1$ (with probability $0.75$) or $y_2$ (otherwise).
When we unroll to $y_2$ during training, we zero the gradients with respect to $y_1$, the noise schedule, and the embeddings.
During held-out likelihood evaluation, we always unroll to $y_2$.
During sampling, instead of solving the recurrence from scratch at each step of the diffusion chain, we compute $\hat{x}_\theta'(z_1, 0)$ for the first step and $\hat{x}_\theta'(z_t, \hat{x}_\theta'(z_{t + (1/T)}, \ldots))$ for subsequent steps.

\subsection{Other details}
\input{sections/method_other_details}


%% file: sections/method_other_details.tex
We perform all forward and backward computations in double precision except for the Transformer layers themselves, which happen in \texttt{bfloat16} mixed precision. This comes at a negligible extra cost since the Transformer layers dominate the overall cost.

\paragraph{Architecture choices}
We condition $\hat{x}_\theta(z_t)$ on the timestep $t$ by adding a sinusoidal encoding of $t$ to the Transformer's residual stream before the first layer.
Before feeding $z_t$ into the Transformer, we rescale it by a factor of $\sqrt{1 + \sigma^2(t)}$ which makes each input dimension approximately unit-variance.
Whereas autoregressive Transformers are relatively insensitive to aspect ratio~\cite{Kaplan20}, we find that Plaid performance increases significantly with Transformer depth up to about 16 layers.
We also find that performance is sensitive to the choice of embedding dimension, with small values performing best. In all experiments, we use embedding dimension $16$.

\paragraph{Stochastic sequence length}
Unlike autoregressive models, diffusion language models can only operate on sequences of exactly the same length as those seen during training.
To enable our model to generalize to shorter sequence lengths, we truncate a small random subset of examples seen during training to random lengths.
We observe that truncating even 3\% of examples allows the model to generalize well across lengths without impacting full-length performance.
Short-sequence performance does not improve substantially as we increase the number of truncated examples.

%% file: sections/ablations.tex
\begin{table}
  \caption{Compute-matched ablations of algorithmic components on OpenWebText2.}
  \label{tab:ablations}
  \centering
  \begin{tabular}{lr}
    \toprule
    & NLL bound (val.) \\
    \midrule
    Our full method & $\mathbf{3.89}$ \\
    Our full method ($0.5\times$ compute) & $4.01$ \\
    \midrule
    No learned noise schedule & $4.17$ \\
    No learned embeddings & $4.54$ \\
    No categorical reparameterization & $4.25$ \\
    No output prior & $3.95$ \\
    No learned conditional likelihood & $4.03$ \\
    No self-conditioning & $3.98$ \\
    \midrule
    CDCD~\cite{Dieleman22} (our reimplementation) & $4.23$ \\
    \bottomrule
  \end{tabular}
\end{table}

\section{Ablation experiments}

In this section, we validate different aspects of the Plaid framework through compute-matched ablation experiments.

\subsection{Validating likelihood-based training}

We take a likelihood-based approach in this work for multiple reasons: it has a principled interpretation, it simplifies training and evaluation, and it has yielded strong results in autoregressive models.
Here, we validate that the log-likelihood objective can attain competitive sample quality through a human evaluation.

In diffusion models, the log-likelihood bound is an expectation over noise levels of a reconstruction loss weighted by a specific function of the noise level.
In contrast, most prior work on diffusion models for language~\cite{Li22,strudel2022self} as well as images~\cite{Ho20,Dhariwal21} use heuristic weight schedules.
Motivated by the intuition that human perception is more sensitive to coarse structure than fine details, these typically assign more weight to higher noise levels than the likelihood weight schedule.

We train three Plaid models: one with the likelihood weight schedule (``VLB'') and two with heuristic weight schedules (``Schedule A'' and ``Schedule B'') which we plot in ~\Cref{app:weight_schedules}.
The models are trained on a large dataset of short children's stories which we constructed by finetuning GPT-J~\cite{Wang21*b} on ROCStories~\cite{Mostafazad16}.
Because learning embeddings is only straightforward when training against the likelihood bound, all models use fixed embeddings obtained from a previously-trained known-good model.

We repeatedly asked crowdworkers to choose from a pair of model samples, where one sample came from the likelihood-trained model and the other came from a heuristically-trained model.
On average, crowdworkers preferred the likelihood-trained model over both alternatives: Weighting A's win rate was $0.449$ ($p=0.001$, 95\% CI $[0.417, 0.482]$) and Weighting B's win rate was $0.457$ ($p=0.005$, 95\% CI $[0.425, 0.490]$).
Because we only consider two alternative weight schedules, we do not claim that the likelihood objective yields optimal sample quality, but our results suggest that it performs at least comparably to other choices.

\subsection{Validating algorithmic components}

Having validated our likelihood-based approach, we show in this section that each of the algorithmic components described in \Cref{sec:method} lead to improved likelihoods in a compute-matched ablation study.

We train Plaid models on OpenWebText2 \cite{pile} and report their log-likelihood bounds on held-out data in \Cref{tab:ablations}.
Our reference model (``full method'') is a $16\times384$ Transformer with $28\mathrm{M}$ non-embedding parameters, trained for $92\mathrm{K}$ steps at batch size $256$ and sequence length $256$, corresponding to $1.12 \times 10^{18}$ non-embedding FLOPs. 
For each ablation model, we stay as close to this configuration as possible while preserving the number of non-embedding FLOPs (we exclude FLOPs from the embedding and output projections because these become negligible at large scale).
We observe that ablating each of the components described in \Cref{sec:method} results in a worse log-likelihood.
As a comparison point, we also train a model at half the compute budget ($5.6 \times 10^{17}$ FLOPs) by halving the model size.
See \Cref{app:experiment_details} for more training details.

Finally, as a comparison to prior work, we reimplement CDCD \cite{Dieleman22}, train it following the same configuration, and report its log-likelihood.
We follow the authors' description as faithfully as possible except for the noise schedule endpoints, embedding dimension, and embedding weight initialization, which we tune to maximize log-likelihood.
We observe in \Cref{tab:ablations} that even the half-compute-budget version of Plaid surpasses our CDCD implementation in likelihood.
Note that CDCD was not developed as a likelihood-based model, and the lack of a public implementation means that there are most likely differences between our implementation and the original.


%% file: sections/scaling.tex
\section{Scaling laws for Plaid}

\begin{figure}
\begin{minipage}[t]{0.45\linewidth}
\centering
\includegraphics[width=2.5in]{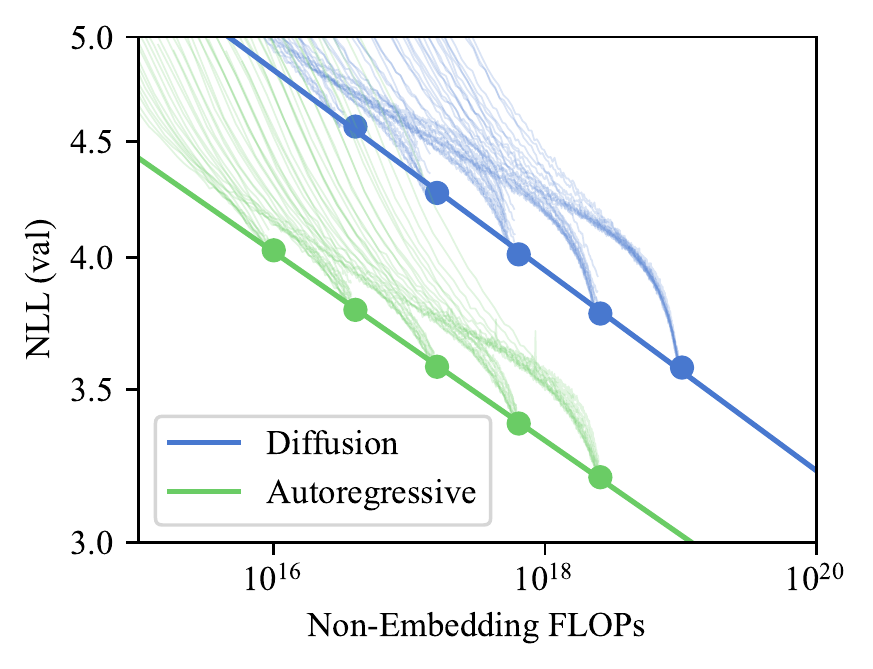}
\caption{Plaid models improve with compute at a similar rate to autoregressive models, but Plaid is less efficient by a constant factor of $64\times$.}
\label{fig:loss_scaling_law}
\end{minipage}
\hspace{0.5cm}
\begin{minipage}[t]{0.45\linewidth}
\centering
\includegraphics[width=2.5in]{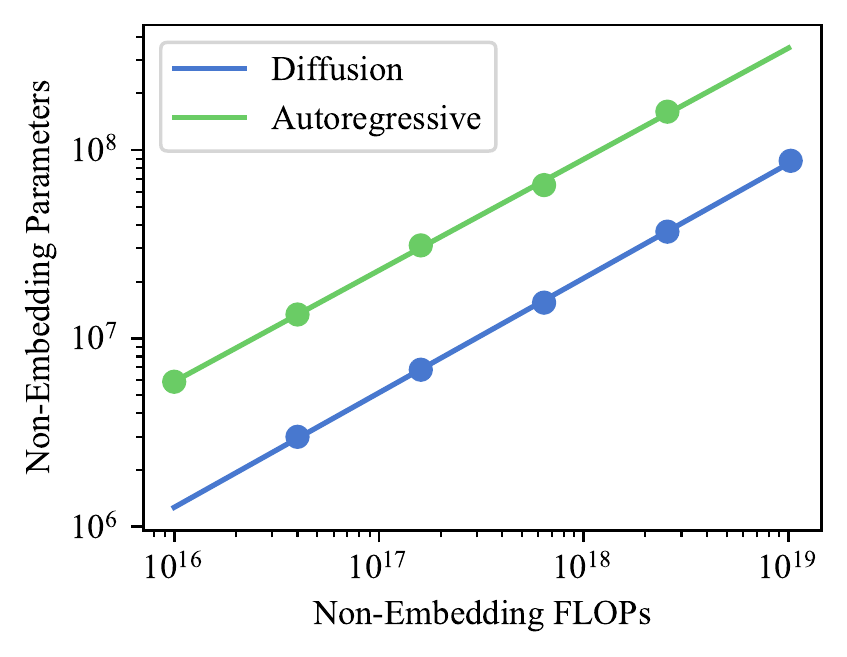}
\caption{Compute-optimal Plaid models should be $4\times$ smaller (and trained for $4\times$ longer) than compute-optimal autoregressive models.}
\label{fig:parameter_scaling_law}
\end{minipage}
\end{figure}

Having developed an algorithmic framework for diffusion language models, we now study its scaling properties in order to guide large-scale training of Plaid models.
In the case of autoregressive models, the work of \citet{Kaplan20} demonstrates that model log-likelihoods follow a log-linear \textit{scaling law}: across many orders of magnitude, training with more compute predictably improves likelihood.
Using these scaling laws, \citet{Kaplan20} and \citet{Hoffmann22} accurately predict the optimal model size as a function of the given compute budget across many orders of magnitude.
Both results together enable effective large-scale training.
In this section, we experimentally determine them for Plaid models.

\subsection{Methodology}

Our main experimental method will be an IsoFLOP analysis \cite{Hoffmann22}.
We first fix a set of FLOP budgets $\{C_1, \ldots, C_K\}$.
For each budget $C$, we train models with different sizes $\{N_{C,1}, \ldots, N_{C,M}\}$ and perform a quadratic fit of the loss $L$ to $\log N$.
We plot all the data along with quadratic fits in \Cref{app:isoflop_profiles} and find that the fits approximate the data well.

The minimum of the quadratic gives us the compute-optimal loss $L^*_C$ and corresponding model size $N^*_C$ for that budget.

Given the compute-optimal loss $L^*_{C_i}$ for each FLOP budget $C_i$, we fit the parameters of a \textit{loss scaling law}
\[ \min_{\alpha, \beta}\sum_i (\log(L^*_{C_i}) - \beta\log(C_i) - \alpha)^2\]
which can then be used to predict the compute-optimal loss as $L^*(C) = \alpha C^\beta$.
We also fit a \textit{parameter scaling law} $N^*(C)$ in the same fashion from the model sizes $N^*_{C_i}$.

We perform IsoFLOP analyses for both Plaid and autoregressive models in order to compare the results.
We choose compute budgets log-uniformly between $10^{16}$ and $10^{19}$ FLOPs and corresponding model sizes heuristically.
We choose learning rates using $\mu$Transfer~\cite{yang2022tensor} and batch sizes, weight decays, and aspect ratios by well-tuned heuristics.
When computing FLOPs, we exclude FLOPs from the embedding layers and output projections.
This enables us to use much smaller compute budgets than \citet{Hoffmann22}, but it causes our autoregressive scaling law to differ slightly from theirs.
We consider this acceptable since we are mainly interested in the differences between our autoregressive and Plaid scaling laws.

\subsection{Loss improves predictably with compute}

We plot both of our scaling laws in \Cref{fig:loss_scaling_law}.
Our first finding is that over many orders of magnitude, the compute-optimal log-likelihood of Plaid models closely matches a power law function of the compute.
Surprisingly, we find that the slopes of both the autoregressive and diffusion scaling laws are almost exactly the same.
These results validate Plaid's scalability and suggest that we can obtain strong improvements by training at larger scale.

Regardless of scale, Plaid models require a constant factor of about $64\times$ more compute to match their autoregressive equivalents.
While this factor is large, our work represents the very first attempt at efficient diffusion model training and focused engineering effort on constant-factor improvements to diffusion models may enable them to perform similarly to autoregressive models in the future. 

\subsection{Compute-optimal training recipe}

Our next goal is to understand how to optimally use a given compute budget $C$ to maximize the held-out likelihood of a model.
Specifically, we must choose between training a large model for fewer iterations or training a small model for longer.
For this, we leverage our parameter scaling law $N^*(C)$ which predicts the optimal model size given a compute budget.

We plot both of our parameter scaling laws in \Cref{fig:parameter_scaling_law} and again find that the trends have nearly the same slope but differ by a constant factor.
Specifically, compute-optimal Plaid models should be about $4\times$ smaller (and therefore trained for $4\times$ longer) than compute-optimal autoregressive models.
The large gap in compute-optimal settings suggests that selecting model sizes based on existing scaling laws \cite{Kaplan20,Hoffmann22}, which were developed for autoregressive models, could incur a substantial loss in the effective compute budget.

%% file: sections/plaid1b.tex
\section{Plaid 1B}

\begin{table}
  \caption{Plaid 1B outperforms GPT-2 124M in zero-shot likelihood across six benchmark datasets from \citet{Radford19}. 
  Our GPT-2 numbers differ from the originals due to striding and detokenization (see \Cref{sec:likelihood_evaluation}).}
  \label{tab:likelihood}
  \centering
  \begin{tabular}{lcccccc}
    \toprule
    & PTB & enwik8 & text8 & WikiText2 & WikiText103 & 1BW \\
    & \scriptsize{(PPL)} & \scriptsize{(BPC)} & \scriptsize{(BPC)} & \scriptsize{(PPL)} & \scriptsize{(PPL)} & \scriptsize{(PPL)} \\
    \midrule
    Plaid 1B (ours) & 74.33 & 1.18 & 1.12 & 29.42 & 28.28 & 77.64 \\
    \midrule
    GPT-2 124M & 87.97 & 1.24 & 1.22 & 35.01 & 35.92 & 87.85 \\
    GPT-2 345M & 64.92 & 1.09 & 1.11 & 26.80 & 26.13 & 67.34 \\
    GPT-2 762M & 53.42 & 1.04 & 1.06 & 23.30 & 22.24 & 59.48 \\
    GPT-2 1.5B & 47.59 & 1.00 & 1.02 & 21.33 & 20.13 & 54.09 \\
    \bottomrule
  \end{tabular}
\end{table}

\begin{table}
  \caption{Chosen unconditional samples from Plaid 1B demonstrate fluent syntax and long-range coherence. See \Cref{app:plaid1b_samples} for un-picked random samples.}
  \label{tab:unconditional_samples}
  \centering
  \begin{tabular}{p{6.53cm}p{6.53cm}}
    \toprule
\scriptsize{\texttt{New research rolled out at an annual scientist meeting finds that the industry will need to recover between 4,000 and 10,000 tons every year of fracked and produced oil and gas fossil reserves in order to do that, according to an analysis done by lead author Dr. Ernesto Monteiro of the University of Alberta in Canada.\newlinesymbol{}\newlinesymbol{}The eye-watering figure represents the total amount of oil and gas — shale and natural gas produced, extracted and sold — will likely need to be recovered in coming years to meet the carbon mitigation goals.}~\textcolor{mediumblue}{\nohyphens{...[698~words~omitted]...}}~\texttt{the team concluded in a report published in an academic journal prepared for the annual meeting of the National Academies of Sciences.}} &
\scriptsize{\texttt{The Barcelona Golf Course doesn’t look like a golf course, but it is an oasis for gardening in this busy city.\newlinesymbol{}\newlinesymbol{}It’s a massive course spread over 120 acres with fairways that are split right down the middle, unlike the designs on most golf courses. The course enjoys the stunning view of the skyline above it.\newlinesymbol{}\newlinesymbol{}A giant oak tree almost 40 meters in diameter serves as one main highlight to the golf course’s design.}~\textcolor{mediumblue}{...[414~words~omitted]...}~\texttt{The new golf course is accessible on the area’s busy streets with shops and restaurants, so the community can enjoy all the leisure activities in the green space.\newlinesymbol{}\newlinesymbol{}The team uprooted the previously existing Pérez Tree, to make room for the new trees to complement Gillet Park.}}\\
    \bottomrule
  \end{tabular}
\end{table}

\begin{table}
  \caption{Chosen conditional samples from Plaid 1B in different zero-shot control settings. Highlighted spans are prompts. See \Cref{app:plaid1b_samples} for un-picked random samples.}
  \label{tab:conditional_samples}
  \centering
  \begin{tabular}{p{13.5cm}}
    \toprule
    \small{\textbf{Prefix completion:}}\\
    \scriptsize{\texttt{\hl{Generative models of text are very versatile: they can be used} as a data classification model and also incorporated into multiple data processing engines.\newlinesymbol{}\newlinesymbol{}In this article, we present two new neural memory models capable of processing terabytes of data and the neural networks and computational techniques that are used in those models.}}\\
    \small{\textbf{Infilling:}}\\
    \scriptsize{\texttt{\hl{A year ago in Paris,} prior to the tournament, I went to Elijah's to eat and get drunk. Everyone in the venue was seventeen. I was there for a few minutes and then I went back to the event. \hl{Wow, what a great day!} So relaxed and too happy. I do not think I was always like that.}}\\
    \small{\textbf{Token-level weights (5$\times$ weight on ``law''):}}\\
    \scriptsize{\texttt{\hl{Let's talk about law and medicine.}\newlinesymbol{}\newlinesymbol{}\newlinesymbol{}\newlinesymbol{}In her dissent, Justice Ron Sen, a veteran administrative law judge, points out that the decision "ignores the fact that the original separation agreement was reached by binding arbitration" that responded to "the legitimate ethical concerns of the university administration," which is what lies "at the heart of law and medicine."}}\\
    \small{\textbf{Token-level weights (5$\times$ weight on ``medicine''):}}\\
    \scriptsize{\texttt{\hl{Let's talk about law and medicine.}\newlinesymbol{}\newlinesymbol{}In part because of advancements in technology, personal information about medical and drug use is spreading. Healthcare professionals across the nation rely on this personal data to make decisions about drug prescriptions and clinical trials and monitor people at immediate risk of serious or chronic diseases.}}\\
    \small{\textbf{Lexical constraints (``Donald'' anywhere):}}\\
    \scriptsize{\texttt{Also facing legal challenges is \hl{Donald} Trump's executive order banning immigration from seven Muslim-majority countries that is facing a temporary halt, with nothing scheduled to go into effect. Two federal judges have ruled that such an order violates the establishment clause.}}\\
    \small{\textbf{Composition and negation (``Donald'' anywhere and ``Trump'' nowhere):}}\\
    \scriptsize{\texttt{A month later, with little time to spare, the government hired \hl{Donald} V. Davis, a former senior aide to Senator Tom Mondale of Minnesota and former Chief Security Operations Officer at the White House, to lead tactical centers.}}\\
    \bottomrule
  \end{tabular}
\end{table}

To demonstrate the scalability of Plaid models and achieve our goal of outperforming an autoregressive model in likelihoods, we train, evaluate, and release a large Plaid model called Plaid 1B.
Plaid 1B is a Transformer-based Plaid model with 1.3B parameters, trained for 314B tokens on OpenWebText2~\cite{pile}.
In total, Plaid 1B was trained for $2.5 \times 10^{21}$ FLOPs, which to our knowledge equals the largest purely diffusion-based language model trained in prior work~\cite{Dieleman22}.
We give further training details in \Cref{app:experiment_details}.

\subsection{Likelihood evaluation}
\label{sec:likelihood_evaluation}
We evaluate Plaid 1B's likelihood in a zero-shot setting on a suite of six benchmark datasets originally used in \citet{Radford19}: Penn Treebank~\cite{Marcus99}, enwik8 and text8~\cite{hutterprize}, WikiText2 and WikiText103~\cite{Merity16}, and the One Billion Word corpus~\cite{Chelba13}.

\citet{Radford19} use sliding windows of size 32 in their likelihood computation.
As a non-autoregressive model, Plaid doesn't support sliding-window likelihood evaluations, so we use non-overlapping 1024-token sequences\footnote{We choose the splitting boundaries using the Plaid tokenizer, yielding sequences slightly shorter than 1024 tokens under the GPT-2 tokenizer.} when computing likelihoods.
Following \citet{Radford19}, we use heuristic invertible detokenizers for PTB, 1BW, and WikiText to minimize the effect of tokenization artifacts on the perplexity results.
For a fair comparison, we also recompute GPT-2 likelihoods using the same protocol, resulting in different numbers than \citet{Radford19}.

In~\Cref{tab:likelihood} we observe that Plaid 1B consistently outperforms the 124M parameter GPT-2 model, demonstrating that diffusion models are capable of scaling to perplexities on par with a small modern autoregressive model.

\subsection{Unconditional samples}

We generate from Plaid 1B by starting from $z_1 \sim \mathcal{N}(0, \sigma^2(1)I)$, performing ancestral sampling of $p(z_{t-(1/T)}|z_t)$ for $T=4096$ steps, and finally $\arg \max p_\theta(x|z_0)$.
Following~\citet{Dieleman22}, we sample using a \textit{score temperature} of $\tau=0.9$, which in our formulation corresponds to adding $\frac{1 - \tau}{\tau}(\hat{x}_\theta(z_t) - z_t)$ to $\hat{x}_\theta(z_t)$ at each step.

We generate unconditional samples with sequence length 1024 and present chosen samples in \Cref{tab:unconditional_samples}.
We observe that the model is capable of generating fluent text and remaining on-topic over several hundred words.
We provide random un-picked samples in \Cref{app:plaid1b_samples}.

\subsection{Zero-shot control}

Although Plaid models are trained in a purely unconditional fashion, we present a zero-shot control technique called \textit{token guidance} which allows us to implement a number of conditioning structures at generation time.
We begin with \textit{classifier guidance}, a technique which allows diffusion models to generate samples conditioned on an arbitrary attribute $y$.
Classifier guidance first trains a probabilistic classifier $p(y|z_t)$ of $y$ given noisy latents $z_t$, and then biases the diffusion model's sampling steps by a \textit{guidance term} derived from the gradient of the classifier probability $\nabla_{z_t} \log p(y|z_t)$.
Now, recall from \Cref{sec:categorical_reparameterization} that our denoiser $\hat{x}_\theta(z_t)$ is parameterized in terms of a model $f_\theta(z_t)$ which learns the distribution over the token $x^{(i)}$ at each position $i$ given $z_t$.
We can therefore implement many different conditioning structures via classifier guidance on probabilities derived from $f_\theta$ itself.
We give a few examples:

\textbf{Conditioning on a span:} We perform guidance on the joint probability of the span under the factorial model $p(x^{(a:b)}|z_t) \propto \prod_{i=a}^{b} p(x^{(i)}|z_t)$, where $f_\theta$ estimates each factor in the product. This lets us implement prefix completion and infilling as special cases.
\textbf{Lexical constraints:} In order to condition on the presence of a token without specifying its location, we perform guidance on the token's probability under the unigram distribution $p(x^{\mathrm{(any)}}|z_t) \propto \sum_i p(x^{(i)}|z_t)$, where $f_\theta$ estimates each term in the sum.
\textbf{Token-level weights:} We can emphasize a specific conditioning token by multiplying the corresponding guidance term by a scalar weight.
\textbf{Negation:} We condition on the negation of an attribute $y$ by performing guidance on the complement probability $1-p(y|z_t)$.

Using Plaid 1B and token guidance, we generate samples under various zero-shot control settings.
We present chosen samples in \Cref{tab:conditional_samples} and random samples in \Cref{app:plaid1b_samples}.
Despite being trained unconditionally, Plaid 1B is able to follow diverse conditioning structures.


%% file: sections/related.tex
\section{Related work}
\label{sec:related_work}

We contribute to a growing body of work on diffusion-based language models~\cite{Li22,Chen22,gong2022diffuseq,strudel2022self,han2022ssd,Dieleman22,gao2022difformer,lovelace2022latent,yuan2022seqdiffuseq,lin2022genie,ye2023dinoiser,han2023ssd2}.
Our biggest departure from those works is that we aim for strong likelihood performance, which to our knowledge has not been attempted in any prior work except for an appendix result from \citet{Li22}.
We therefore benchmark against well-known autoregressive models instead of prior diffusion language models.

The work most comparable to ours is CDCD \cite{Dieleman22}, which is also a strong general-purpose diffusion language model.
However, without the ability to use standard likelihood-based benchmarks~\cite{Marcus99,hutterprize,Merity16}, it is difficult to say precisely where CDCD stands in comparison to autoregressive models: in every result, either CDCD underperforms the autoregressive baseline, or the evaluation metric saturates and lacks the statistical power to distinguish the models.
Many of the other works above share similar difficulties.
In contrast, our likelihood-based approach enables unambiguous comparisons to widely-known models.

Other diffusion language model works consider more constrained settings like controllable generation \cite{Li22} or sequence-to-sequence tasks \cite{Chen22,gong2022diffuseq,gao2022difformer,yuan2022seqdiffuseq,lin2022genie,ye2023dinoiser}, or propose hybrid approaches involving pretrained autoregressive models \cite{lovelace2022latent, han2023ssd2}. Particularly, in concurrent work, \citet{han2023ssd2} finetune an OPT 13B~\cite{zhang2022opt} model into a hybrid model which is autoregressive over 25-token blocks and uses diffusion within blocks. Compared to their work, we focus on the more general setting of training a fully diffusion-based language model from scratch.

Finally, our work builds on recent advances in diffusion models for images \cite{SohlDickst15,Ho20,Rombach21,Kingma21,Dhariwal21}.
Most notably, we adopt the framework of Variational Diffusion Models \cite{Kingma21} and extend it to language modeling.
	

%% file: sections/conclusion.tex
\section{Conclusion}
In this work, we have taken the first steps toward a competitive likelihood-based diffusion language model.
We built Plaid 1B, which matches GPT-2 124M in likelihood by combining several algorithmic improvements and a scaling law analysis.
Our ablations show that maximizing likelihood does not substantially harm sample quality, and we show samples are fluent in both unconditional and zero-shot conditional settings.
Despite this progress, substantial work remains: Plaid narrows the compute-efficiency gap between diffusion and autoregressive language models to $64\times$, and we view this gap as a tractable and exciting open problem that may be addressed with further research.

%% file: sections/appendix_weight_schedules.tex
\section{VLB and heuristic weight schedules}
\label{app:weight_schedules}

\begin{figure}[h!]
\centering
\includegraphics[width=3in]{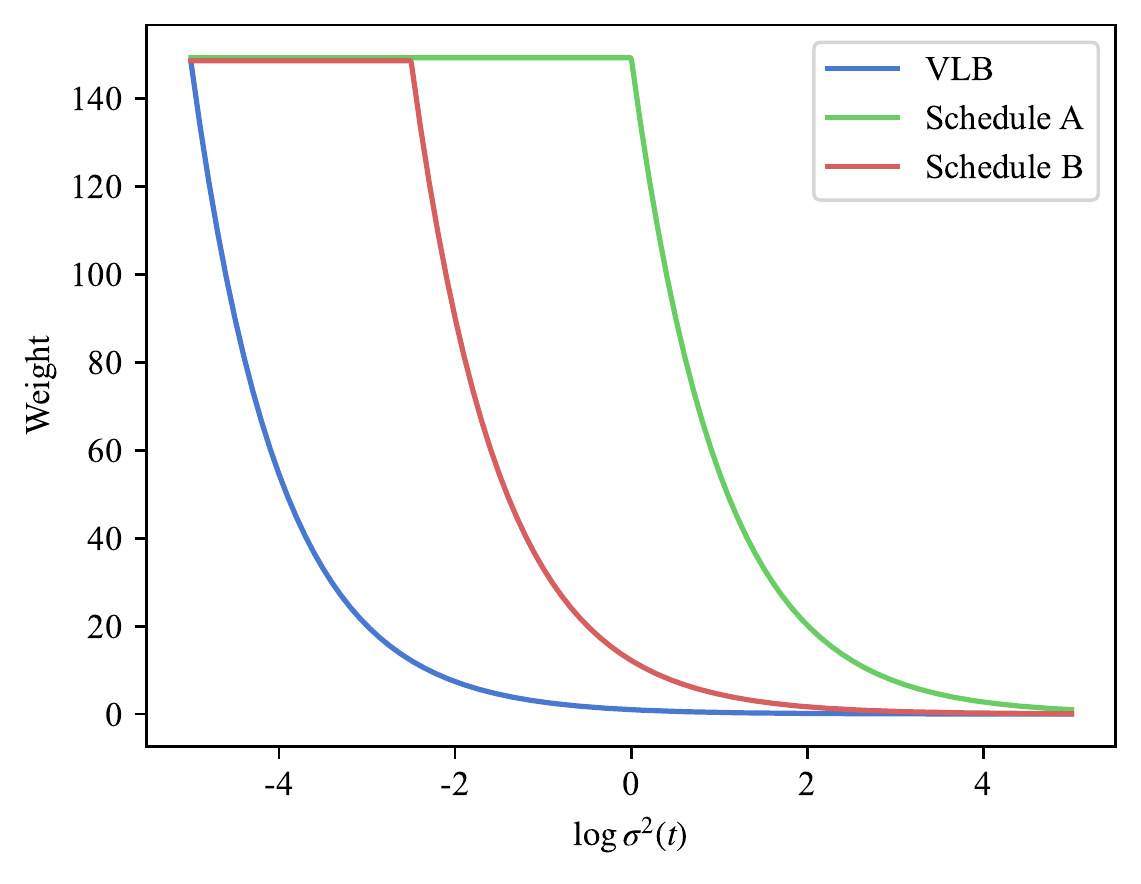}
\caption{VLB weight schedule and and heuristic weight schedules used in ablation experiments.}
\end{figure}

%% file: sections/appendix_experiment_details.tex
\section{Experiment details}
\label{app:experiment_details}

\subsection{Dataset}

Unless otherwise noted, all models in this work are trained on a subset of OpenWebText2 \cite{pile} which we filter to remove documents labeled as non-English.
The data is tokenized using a 32K-token BPE tokenizer which we train on the OpenWebText2 training split.

\subsection{Architecture}

We use standard pre-activation Transformers models with RMSNorm normalization layers and GeLU nonlinearities throughout.
Unless otherwise noted, all Plaid models have 16 Transformer layers, which we found to be approximately optimal for our scale.
We scale autoregressive model depth approximately following \cite{levine2020depth}. 
For efficiency, our implementation uses FlashAttention \cite{dao2022flashattention} and other fused kernels wherever applicable.
We train at sequence length 256 for all experiments except Plaid 1B.

\subsection{Optimization}

We optimize all models using AdamW with parameter-specific learning rates derived by $\mu$Transfer \cite{yang2022tensor} based on a learning rate of $1.4 \times 10^{-3}$ at width 256.
Each parameter's weight decay is set to $\frac{4 \times 10^{-5}}{\eta}$ where $\eta$ is that parameter's learning rate.
We use a linear warmup on the learning rate and weight decay over the first 2500 steps, followed by a linear decay to zero over training.
We train at batch size 256 for algorithm ablations and 128 for scaling law experiments.

\subsection{Plaid 1B Training}

We increase the base $\mu$Transfer learning rate to $2 \times 10^{-3}$ (at width 256).
The denoiser network is a Transformer with 24 layers of width 2048 and a vocabulary size of 32K tokens, for a total of 1.3B parameters.
We train for 1.2M steps at batch size 256 and sequence length 1024, for a total of 314B tokens.
All other details are as written above.

%% file: sections/appendix_isoflop_profiles.tex
\section{IsoFLOP profiles}
\label{app:isoflop_profiles}

\begin{figure}[h!]
\begin{minipage}[b]{0.45\linewidth}
\centering
\includegraphics[width=2.5in]{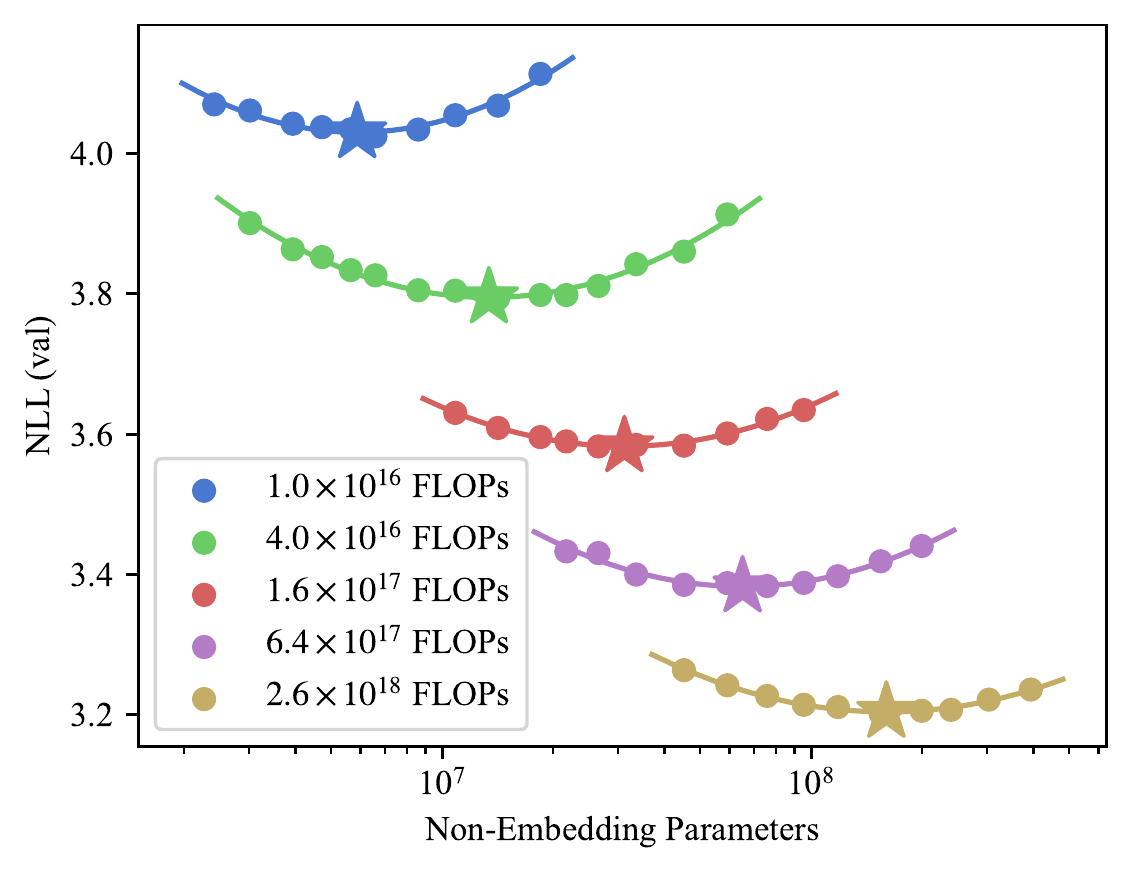}
\end{minipage}
\hspace{0.5cm}
\begin{minipage}[b]{0.45\linewidth}
\centering
\includegraphics[width=2.5in]{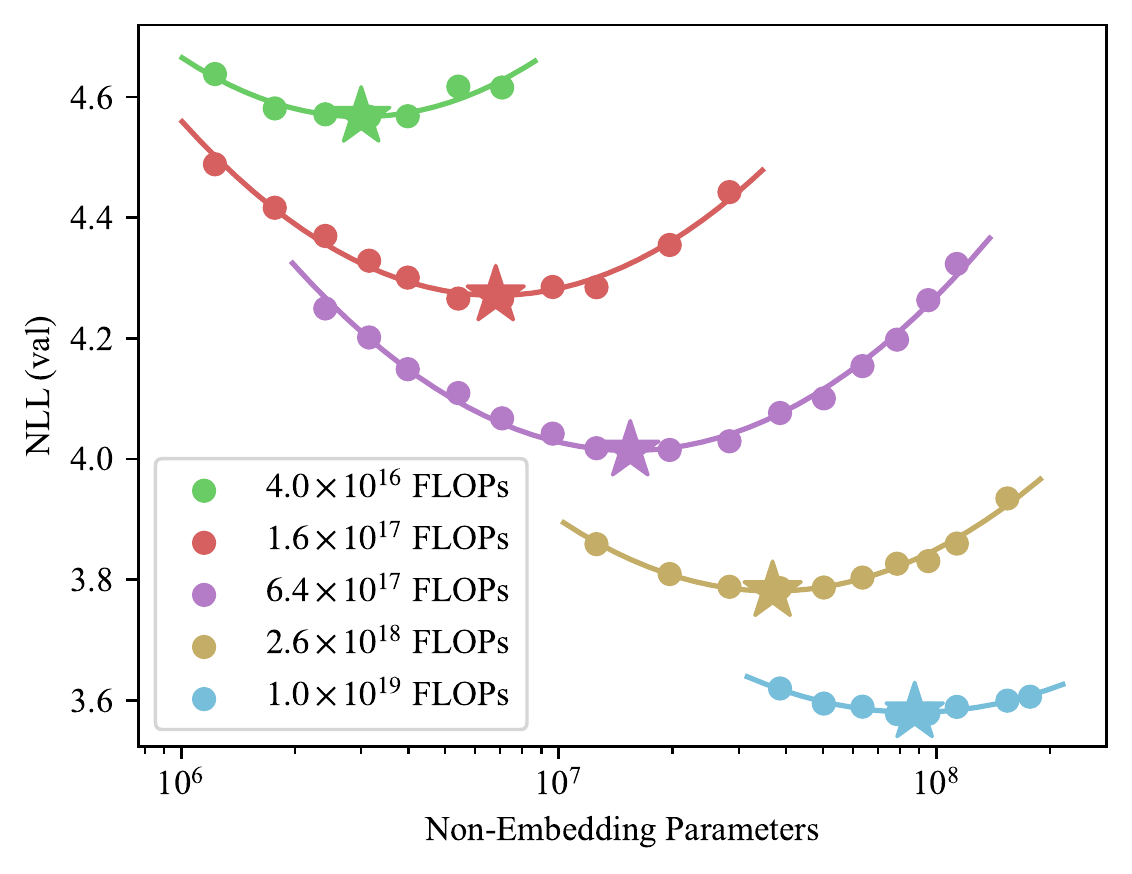}
\end{minipage}
\caption{IsoFLOP profiles for autoregressive models (left) and diffusion models (right).}
\end{figure}

\pagebreak

%% file: sections/appendix_samples.tex
\section{Plaid 1B Random Samples}
\label{app:plaid1b_samples}

\subsection{Unconditional}

\tiny{
\begin{verbatim}
 tends to create a ripple effect in the economy, and economist Kevin Milligan said it’s probably a
 testament to how worn-down the economy and consumers are, particularly at the low end.\n\n“It’s probably due to things
 like investment slowing in the natural resource sector, weakness in stock buybacks in the corporate sector,
 corporatism. And, of course, (we) have to look at things about what the Bank of Canada is doing, the uncertainty on
 the horizon in terms of trade, and the exchange rate, and all of these things tend to pull down the economy,
 particularly the low end.” Tweet This\n\n“I think the economy in general has been feeling the squeeze of low credit
 rates, (and) especially low mortgage rates,” he said.\n\nStory continues below advertisement\n\n“It’s difficult to be
 selling at rates well below their 10-year average, so I think those effects do come back into the economy as well, and
 I’d wonder if it’s pulling down consumption and investment activity.”\n\nConsumers and businesses are really looking
 for more and more type of stimulus action not just from the Bank of Canada, but from the federal government too, he
 added.\n\nHe said they’re not getting a message from Canadians about the state of the economy and he thinks many
 people are wondering if they did get that message from Prime Minister Justin Trudeau in December.\n\n“I’d be happy for
 people to hear a message, and that has to come through,” he said.\n\nDespite the second interest rate hike, Milligan
 said there’s reason to think the odds of an interest rate hike have risen from about one-third to about two-thirds in
 the first part of next year.\n\n“I would, on balance, expect a sustained number of rate hikes – even a third hike.”
 Tweet This\n\nHe said those predictions are becoming more consistent as well, with a reported saying from economist
 Chew Ross this week that most economists consider a fourth rate hike a “near certainty” for 2019.\n\nMilligan said
 he’s optimistic about the economy in Canada.\n\n“I think we’re probably in the middle of seeing a strong bottom and
 picking back up on a strong GDP, and we’re very much into this directional trend for the economy now that we see.”
 Tweet This\n\nStory continues below advertisement\n\n“I think we’re on track for a run of three straight years of
 zero- to one-per cent of GDP growth, which would be quite strong. And we’re not quite back at an extended
 bottom.\n\n“But I think we were probably hitting close to bottom in the recession in 2009, and we’re clearly in that
 situation. So at some point, you want to move up, and of course, one reason would be another surge in the
 economy.”\n\nAsked if the possibility of a fourth rate hike in 2019 could help business sentiment going into the
 federal election, Milligan said heightened expectations are always a good thing.\n\nHe also said there’s more
 opportunity for the Bank of Canada to announce some form of monetary policy stimulus following the second interest
 rate hike at its next policy meeting, and financial markets around the world would likely rise in response to such a
 move.\n\nStory continues below advertisement\n\n“But if they’ve absorbed the downside of those rate hikes in other
 ways, I’m not convinced that they are their biggest concern in terms of their intentions.”\n\nWith files from
 Bloomberg<|endoftext|>A man has been arrested for allegedly crashing into a group of three children walking
 close to Ariana Grande from London’s Electric 6 Club.\n\nThe three three-year-olds and their father were walking to
 the singer’s home from London’s Shelter Art after the show when they came across a black sports car driving at about
 85 mph.\n\nThe unidentified man, who is believed to be the driver of the ‘sorry’, passed them about 15 or 20 metres
 suddenly and one child was caught up in a collision at the rear, the Metropolitan Police said.\n\nThe girl was not hit
 by the car, the Evening Standard reports.\n\nThe driver, arrested on suspicion of driving while working and driving
 recklessly, has been released on court bail without further comment.\n\nFollowing the incident, the girl was rushed to
 hospital with an urgent call for coronary coronary arrest due to severe shock. The child’s parents expressed their
 shock at the “serious” incident, Sky News reports.\n\nJUST IN: ‘The Queen was targeted’: UK Police offer exact timings
 for ‘targeting’ terror attack on the Royal Family https://t.co/gCJaqJnaAi — Breitbart London
 (@BreitbartLondon) February 13, 2019\n\nResponding to the incident, Ariana Grande tweeted:\n\nWas to @Meteze show
 in #London last night! We will not forget those in the early

 +147402] Gibdobber (19.55) [157000+148691] Arch-Princess (19.53) [157000+147720] Snow White (19.42)
  [157000+147790] Magdalene (19.41) [157000+148447] Corrin (18.93) [157000+148068] Pinocchio (18.92)
  [157000+148396] Ayya (18.85) [157000+148384] Alaina (18.84) [157000+148384] Olaf (17.76) [157000+148791] Chu (17.72)
  [157000+148493] Robin (17.62) [157000+147823] Papyrus (17.64) [157000+148396] Elsa (17.64) [157000+148384] Geralt
  (17.62) [157000+148659] Marie (16.52) [157000+148659] Anna (16.52) [157000+148597] Leon (16.52) [157000+148697] Paul
  (16.48) [157000+148697] Reg (16.46) [157000+148697] Shu (16.44) [157000+148659] Aurora (16.43)
  [157000+148597] Heinrich (16.33) [157000+147798] Perraman (16.01) [157000+148696] Peter (15.91)
  [157000+148592] Papyrus (15.72) [157000+148596] Arch-Maid (15.72) [157000+148069] Belldobber (15.66)
  [157000+148396] Corrin (15.64) [157000+148591] Helena (15.60) [157000+148397] Hazel (15.57) [157000+148594] Alva
  (15.56) [157000+148653] Olaf (15.53) [157000+148149] Anna (15.53) [157000+148597] Hans (14.49)
  [157000+148597] Heinrich (14.43) [157000+148657] Aurora (13.42) [157000+148656] Shu (13.38)
  [157000+147823] Arch-Princess (13.329) [157000+147692] Elsa (13.37) [157000+148597] Tamara (13.37)
  [157000+148437] Richard (13.36) [157000+148575] Geralt (13.35) [157000+148144] Reg (13.34) [157000+148256] Arthur
  (13.33) [157000+148659] Papyrus (13.27) [157000+148697] Arthur (13.24) [157000+148109] Peter (13.24)
  [157000+148597] Chu (13.17) [157000+14785] Marie (13.08) [157000+148659] Anna (13.08) [157000+148659] Paul (13.07)
  [157000+148657] Magdalene (13.07) [157000+147780] Geralt (13.07) [157000+147780] Olaf (13.06) [157000+148255] Reg
  (13.05) [157000+148255] Richard (13.04) [157000+148255] Arch-Princess (13.02) [157000+148596] Aurora (13.02)
  [157000+148596] Heinrich (13.01) Perraman (12.82) [157000+148696] Hans (12.70) [157000+148349] Papyrus (

, and occasionally some of our recommendations may have changed. With that said, if you choose to buy something using
  the retail links included in this post, Altdica may earn a small commission. However, all of our product reviews are
  written by independent experts with no relevant agenda to consideration.<|endoftext|>"This is about the
  people," said lead attorney Steven Holstein, a Philadelphia lawyer who represented the company targeted by the state
  robbery task force in the case. "When some nut across the state has the ability to write a constitutional amendment
  and rewrite a constitutional amendment in a way that makes it antithetical to the state of Pennsylvania, and say they
  are against it ... then it doesn't really matter what the people of Pennsylvania want."<|endoftext|>Police
  officers were paid an extra £425 for each working hour in 2019, despite many of the people involved in the
  high-profile Windhouse scandal being involved in front-line duties, a service review has revealed.\n\nThe total pay
  paid in 2019 was a minimum of £42,487, while officers earned a maximum of £38,254 - with an average hourly pay
  increase of £2.46 per hour.\n\nThe salaries amount to an average hourly pay rise of more than £5 between June 2018
  and June 2019.\n\nFinancing of officers' services rose by 7% in the past year over the previous year, the report by
  the Association of National Police Officers (ANPO) revealed. Investment in non-police trusts rose by 17% in the same
  period.\n\nANPO chief executive Brian Kirk said: "Police officers rightly want to be rewarded for their service, so
  they do not have to choose between keeping communities safe and delivering good policing.\n\n"This extra pay enables
  police officers to consider new career opportunities, provide support to those most in need and promote social
  inclusion."\n\nThe Scottish Government said in 2017 the police service had more than 23,000 officers, a third of whom
  were front-line.\n\nThe current pay structure was introduced in the late 1990s to compensate officers in frontline
  commands with high crime rates.\n\nScottish Labour Shadow Justice secretary Michael Russell said the fact officers
  were getting paid adequately showed "something urgently needs to be done".\n\nHe said: "It's completely bonkers that
  taxpayers would expect to cough up an extra £425 per hour for a police officer's shift, especially when frontline
  staff like health workers, social services, fire and ambulance continue to face regular annual cuts.\n\n"I can hardly
  believe ministers continue to grant hugely lucrative contracts, while freezing its core wages, to a health service
  which is already struggling as a result of huge budget cuts, and which the SNP has consistently opened up as a no-go
  zone over the last four years.\n\n"Why have ministers allowed higher officer salaries in the police while they have
  delivered a decade of devastating budget cuts to the NHS?"\n\nScottish Liberal Democrat leader Willie Rennie
  said: "It is notable that there has been a significant improvement to current pay arrangements in relation to cover
  officer roles.\n\n"However, any rise in pay for the workforce would also need to be brought in line with the huge
  cuts we have already seen in frontline services.\n\n"The Scottish Government showed how they last week awarded
  zero-cost contracts for labour companies without even looking at what the work cost, and the clear message is that
  they are keen on making frontline organisations look more modest rather than worrying about budget cuts.\n\n"There
  will be more of the usual bonuses coming before the Government shows how finance and good colleagues will be putting
  a concerted effort together to find some resources to offset some of these increased costs, instead of deciding the
  way to save the workforce is a cutters field."\n\nNicola Rowland, the general secretary of the RNQ, said: "Front line
  firefighters and police officers are asked to endure some of the most distressing things on a daily basis."\n\nThe
  average police officer pay rose to £43,085 in 2019, after a two-year increase, the ANPO said.\n\nThe rise in law
  enforcement is seen to have been caused by an ageing population and the overseas population and the deepening of
  recent job cuts in the Scottish public sector.<|endoftext|>PASADENA, Calif. (KABC) -- A 17-month-old girl
  has been returned to the Potnona Police Department Monday afternoon. She was taken from a home in the 3400 block of
  Ponderosa Avenue near Fairfax Boulevard and Lelgar Street.Police say it is believed that the child, who is staying
  with her family, was taken from her home between July 20 to July 22.Forensic testing will likely be performed later
  this week.<|endoftext|>It was an odd fun fact that made many left-wing activists vehemently oppose the
  nomination of Brett Kavanaugh to the U.S. Supreme Court. On his yearbook page was a subtote entitled “Don’t Let Them
  F***** Your Body Forever,” a reference to the oral contrace

 of carbon emissions needed to meet the 2 degree Celsius target set forth in the last assessment of the
 Intergovernmental Panel on Climate Change (IPCC) in 2013. New research rolled out at an annual scientist meeting finds
 that the industry will need to recover between 4,000 and 10,000 tons every year of fracked and produced oil and gas
 fossil reserves in order to do that, according to an analysis done by lead author Dr. Ernesto Monteiro of the
 University of Alberta in Canada.\n\nThe eye-watering figure represents the total amount of oil and gas — shale and
 natural gas produced, extracted and sold — will likely need to be recovered in coming years to meet the carbon
 mitigation goals. The amount where between 4,000 and 10,000 metric tons of reserves for every 10 metric tons of carbon
 dioxide they displaced.\n\n“The benefits of carbon capture increase with the amount being injected,” Monteiro wrote in
 an article accepted to publish in the Proceedings of the National Academy of Sciences which DeSmog obtained last
 week.\n\nWhile the theoretical ability to limit 2 C warming with fossil fuel carbon budgets is obvious, such action is
 not taken by the oil and gas industry. Industry groups led by RRC openly seek to capture 38 million tons of carbon
 dioxide emissions annually from operating reserves and transfer a portion of that back to the oil industry as
 revenues. “Comparing natural gas reserves under production with proven oil reserves remaining reveals one potential
 implication, which is the additional natural gas production that would be required to fully satisfy the 2100 oil
 supply balance,” the paper explained. By Monteiro’s estimates, natural gas production will grow by the equivalent of
 about 4 Gb per year by 2100, while at the same time oil production will decline at about 1 Gb per year, or a total of
 2.66 Gb per year in balance. If all the resources at hand could be extracted, of which no amount is available at all
 times, it would be practical to use all of that fossil fuel along with existing operating wells. But as natural gas
 production rises, new wells could replace declining oil production with an equivalent number of fossil fuels to
 reserves.\n\nCarbon reductions at any level involve “costs” — additional expenditures that necessarily arise from oil
 and gas extraction and associated energy production, including physical production, liquefaction capability, handling
 and shipping fees. The research found oil and gas companies bid to recover as much natural gas as possible at all
 early stages of exploration and production. Natural gas, while more profitable, is much riskier than coal. The paper
 explains: “Extracting and producing natural gas exploits the time, location and physical composition of the fossil
 record. Each well represents a different geological formation with different natural gas characteristics. These
 characteristics mean that the amount of natural gas recovered from a well is not determinative of the productivity of
 the well from which it is extracted. Reference reserves provide an indication of the natural gas resources that may
 exist in different shale formations and the potential of these resources based on past and future discoveries.”\n\nThe
 only kind of geomorphological formation that can be considered very promising for oil and gas production are mesopolar
 formations, those within one mile from surface to three miles below ground. Mesopolar formations are generally
 regarded as more inflexible than permafrost or acid shale formations, less stable than other subsurface formations or
 fracked tight rock, more susceptible to seepage, and more susceptible to oil production costs. For these reasons, the
 reserve estimate put out at the annual meeting does not consider natural gas production to be an economic viability
 from existing mesopoint formations.\n\nThe net result is that the industry will lose out on billions of dollars in
 revenues to pay part of the cost of carbon sequestering, underground scrubber systems and other forms of carbon
 capture from any new natural gas reserves of coming years. In order to fuel plants, the primary gas is oxygen; when
 undertaking oil and gas production and development that oxygen is extracted from the air and converted into methane
 that is vented through rock. It then is dispersed at the surface using various methods, such as proximity to carbon
 sequestration facilities and underground scrubber systems, to provide the chance to store carbon dioxide that would
 otherwise have been leaked back into the atmosphere.\n\nIn 2014, a team of scientists from the National Academies of
 Sciences calculated that deferring as many as 3000 metric tons of carbon dioxide emitted from oil and gas reserves
 already on Earth to new ones would come at a cost of 5 cents annually to the oil industry per ton. Revenues would then
 be about \$35-60 billion. “Financials indicate that nonconventional expenditures, including relatively large ones,
 will have a material net impact on industry and substantial constrains on operating income and free cash flow,” the
 team concluded in a report published in an academic journal prepared for the annual meeting of the National Academies
 of Sciences. It concluded that the finance companies that actually receive profits from the stranded reserves, which
 offer little more than \$4

] have the Senate convict a sitting president.”\n\nOn Wednesday, after the House Judiciary Committee drew up articles of
impeachment, McConnell announced that “we have the votes to convict the president.”\n\nTrump also held a rally, in
which he decried the process as “an illegal coup against me,” and announced that he would “appeal.” The rally seemed to
show that the fight wasn’t over.\n\nWATCH:\n\n“This is not a tragedy,” McConnell said in a statement at the time. “I am
convinced, with every fiber of my being, that the President is innocent. Having solicited the articles of impeachment
from the House, the Senate looks forward to hearing from House managers on the charges and then moving forward to
presenting a case for trial by the full Senate.”\n\nTrump’s GOP rivals and independent senators like Sen. Lisa
Murkowski (R-AK) have publicly supported impeaching the president, while other GOP senators, like Sen. Susan Collins
(R-ME) have openly said they could resign.\n\nThe first major impeachment defection came on Sunday when longshot Rep.
Justin Amash (R-MI) announced his intention to convict. In an email to his colleagues, Amash Jr., who criticized
Trump’s actions in the months leading to and through the impeachment, said he was voting to convict.\n\n“From where I
sit, the evidence is overwhelming. As I am a member of the Judiciary and Intelligence Committees, I have a duty to the
American people, as well as an obligation to my constituents and all who sent me to office, to vote convict,” Amash
said.\n\n<|endoftext|>The Barcelona Golf Course doesn’t look like a golf course, but it is an oasis for
gardening in this busy city.\n\nIt’s a massive course spread over 120 acres with fairways that are split right down the
middle, unlike the designs on most golf courses. The course enjoys the stunning view of the skyline above it.\n\nA
giant oak tree almost 40 meters in diameter serves as one main highlight to the golf course’s design. The course’s
6,486 holes - some six times downhill - are a real treat for newbies and even seasoned golfers.\n\nArcadia for 120
acres was built around Gillet Park Golf Club, an exclusive private mixed-use development. Vergara Andreana’s team
served as the architect for the golf course, with the course designed by 110 Architecture, the local firm working
exclusively with Barca, the city’s Professional Football club.\n\nEven though Andreana is the lead architect on the
project, the team was tasked with creating other lifelines around the course as well as the concrete facade. Aside from
the course, Vergara and Andreana have been involved with at least 20 different, publicly developed projects in
Barcelona.\n\n"Design has to be sustainable"\n\nVergara and Andreana became aware of the Barcelona Golf Course while
participating in an environmental conference. They recalled being surprised by the site’s design, which helped to
heighten their enthusiasm for the project. “Because we are architects, architecture has to be sustainable, either low
energy or low/zero carbon,” Andreana says. “We also want challenges for new projects.”\n\nVergara and Andreana started
in landscape architecture in 1995. At the time, they were involved with relatively small projects located in the
tourism sector. Elsewhere, Vergara and Andreana were searching for a community garden to allow families and people to
experience the magic of nature exploration and gardening.\n\n“We are a architect’s team and we want to improve the
urban environment for our users,” Andreana says.\n\nThe team began to look for a place to design a nine-hole golf
course. In June 2013, they found a spare garden filled with lush grass located in a meadow.\n\nA great opportunity for
diversification\n\nAndreana’s team researched other urban golf courses for the right location. In the end, Gillet Park
Golf Club stood out as the best option. Although the natural gardens needed a restoration for their natural
conservation, the size and location of the golf course offered a unique opportunity to beautify and reinvest in the
urban environment.\n\n“Gillet Park Playground met all the criteria,” Vergara and Andreana explain. “It’s not very
specific, suited the golf course, and had all the right vegetation.”\n\nVergara and Andreana spent just nine months
planting thousands of trees in the western corner of Gillet Park, in the neighborhood of Joan Mariscal. The new golf
course is accessible on the area’s busy streets with shops and restaurants, so the community can enjoy all the leisure
activities in the green space.\n\nThe team uprooted the previously existing Pérez Tree, to make room for the new trees
to complement Gillet Park. It was a quick and convenient solution

 however it requires two-piece manual construction, which the designer does with great skill.\n\nCuboys does great
 detail work, the whole body is mostly to scale, with the obvious exception of the sword guard. His arms and legs are
 also well detailed and the model has pretty good proportions. There are still some details that could probably do with
 a little more refinement that I didn't catch, but in general I like the overall look of this model and it really feels
 realistic.\n\nAll in all I was quite impressed, the model is put together with great materials, good use of miniatures
 and nice detailed details that are really visible.\n\nThis is most definitely another excellent example of modular
 construction in action.<|endoftext|>NEW ROWFORD, Conn. — A man with schizophrenia faces charges for
 punching a 15-year-old boy in the head before using a tire iron to strike the youngster in New Bedford, according to
 police.\n\nJust before midnight Wednesday night, New Bedford police responded to Liberty Mill Avenue and 66th Street
 for reports of a non-consensual felony battery, according to police Lt. Mark Reed.\n\nA child wielding boy, 15, was
 allegedly approached by a man with a Southern accent and mustache, who started to beat the boy before pulling out a
 tire iron, the lieutenant said.\n\nThe suspect continued his assault after the boy's forehead was burned with the
 iron. Reticosis to the boy's face required additional stitches, Reed said.\n\nThe suspect is charged with first-degree
 assault and faces up to five years in prison.\n\n\n\n\n\n<|endoftext|>Today, a massive tree fell off a road
 in the famed resort town of Rosengård, Poland. As of now, there have been no injury reports; however, witnesses told
 Poland's TNT radio station that the trunk of the tree stood at least 3.5 meters tall. It was almost completely
 blocking the road used to enter the popular ski resort town.\n\nAccording to report, the base of the tree has fallen,
 but the abrupt all-round fall still seems a rather odd way for a tree of such nature to fall. Local media speculated
 that a branch had come down, but said that "it's couldn't be ruled out." Miraculously, there have only been a handful
 of injuries, presumably from the rogue trunk.\n\nThe road is still under reconstruction in some areas, and a single
 tree of this size and age toppling on any foundation should not do any real damage. It is reported that the tree will
 be removed and reburied.\n\nUpdate: Local emergency crews are still being dispatched to deal with the situation.
 However, the tree has reportedly now been removed and the road is now open. A supermarket and McDonald's have already
 re-established.\n\nThe base of the tree that fell in Rosengård. The remaining trunk fell and caused no injuries.
 Credit: Peter Owen http://m.flickr.com/photo.php?worldid=126161&category=166902#tif\n\n---\n\nSource\n\nThe Man Who
 Saw Mountains\n\nby Bill Winter\n\nThe resort as was planned. 1978<|endoftext|>Two alleged thefts gained
 entry into a home in the 3300 block of North Monroe Street with guns blazing Monday evening. Residents attempted to
 make entry. Key Details: Two thefts reportedly gained access to the home as residents were outside and were attempting
 to lock their homes. One thepper pulled his gun, out of the house. The other person than reportedly pointed his gun at
 the one man. However, no shots were fired. Police tell 7NEWS the alleged thieves did already know each other. At least
 three other thefts were reported Monday evening. No word yet if these two thefts could be a
 connection.<|endoftext|>Jorge Miles Taleb\n\nDaily Stormer\n\nJuly 8, 2013\n\nThe shooting at a Tucson
 white supremacist rally night is the latest in a string of recent terrorist incidents at white supremacist centers. It
 follows a mass shooting at the Oregon State Capital District in June ending with 9 people in Seattle, Washington, and
 one of two people in Midland, Texas being killed in the attack. After witnessing these tragedies, people have actually
 reacted in a predictable way, and pointed towards the problem of white supremacists and violence being the real
 problem that’s plaguing America.\n\nAccording to reports from the mainstream media, the shooters were not from Oregon
 or Texas, suggesting that they may not have been driven by white supremacists, given the recent history of racially
 motivated attacks while many there were also confused by the motive of the shooters.\n\nThe shooting took place at a
 rally that mentioned the Patriot, but their movement is civil disobedience of the federal government. This of course
 is considered unacceptable and unacceptable by many of those who run the national movement, as it goes against their
 view that the role of government should be eliminated.\n\nThough they do denounce the Occupy movement which is based

\n\nAlexandra Firth reports on the association between higher rates of smoking in the adolescent stages and higher
 obesity in subsequent stages. Like Ridley, Whitfield concludes that this means “even if you’re at the bottom you’re
 one generation away from obesity”.\n\nThis is certainly true, though there are two big problems with it. First, BAME
 adults tend to be smoking at much later ages than those in the adolescent stages, on average at a later age of 34.
 This means that there is a causal relationship between smoking and poverty that is not present when we control for
 poverty or look at controlled factors. The degree of economic segregation, or the factor that *controls* the wealth
 gap – household income according to dollar index – is not nearly as strongly correlated with elevated smoking rates
 among the lowest part of the income distribution (Harrison and Strauss-Daly, 2006, 48-49). A higher part of the income
 distribution tends to have higher smoking rates, so the level of cigarette consumption is in fact correlated with the
 degree of economic mobility (though it’s important to note that it is higher status that has the higher rate of
 shoplifting). On the other hand, economic mobility does not seem to be correlated with the causal relationship between
 smoking in early adulthood and obesity in later adulthood. This means that being at the top of the income distribution
 does not necessarily give someone “higher” status, or more privileged.\n\nSecond, even if you control for
 socioeconomic status, people who are more likely to have higher rates of smoking in early adulthood tend to have to
 higher rates of obesity in later adulthood than those who are less economically privileged, and there’s no way that
 one can infer “teens away” from this, seeing as there is no causal relationship between smoking and mobility, and no
 threshold beyond the top of the income scale that is exhibiting an upward trajectory.\n\nIt is certainly true that a
 greater proportion of BAME adolescents from families with lower incomes are on average less likely to smoke.
 Curiously, Whitfield doesn’t mention that socioeconomic mobility differs significantly between different income
 groups, so this kind of relationship can’t be replicated. There is no causal relation, but speaking, just because
 those at one top of the income scale are more likely to see their kids try cigarettes, does not mean an ongoing
 predisposition towards obesity (or even smoking).\n\nEven though it can’t be proven, and just because it sure seems to
 incite a fairly clear upward trajectory of mobility for people at the top of the income distribution, Whitfield is
 basically saying people who have higher rates of adolescent smoking are poor, and not vice versa. This is another
 misinterpretation/misuse of statistics in public health, and a serious stretch for an economic science that is
 considered relatively bias-free by its supporters and anti-economists alike.\n\nTaking the long view: When statistical
 models have (pertature) interpretations\n\nIn the same interview Whitfield also makes another somewhat questionable
 claim, regarding cigarette smoking and childhood obesity. After finding out that there is no significant causal
 relationship outside of the early stages with the level of smoking, he said:\n\nPeople are now wondering why you would
 try to avoid the early stages if there is so little causal connection. There is always going to be something in there
 about healthy early drinking and preventing an artificial transition.\n\nThe simple fact is that statistical models
 don’t get to consider a person’s personal interpretations. A model that shows there is no adverse relationship between
 smoking and obesity in the young stages does not come close to considering the personal call on the risk of smoking in
 the middle stages; if it is harmful for all, what is the benefit? To understand how interpretations like this are
 sometimes folded into public health policy I think it’s helpful to understand the science. Statistical models do not
 allow people to tease different factors apart, with differentiated trade-offs. Instead, it requires that you take into
 account all of the factors, and come up with the best predicted outcome. Even if one of these models has conflicting
 statistical data, there is no concept of statistical meta-analysis as we have in the sciences.\n\nFor example,
 Whitfield still needs to ask:\n\nif diet is additive, and therefore a lot of people would develop a nicotine habit,
 then, why not just go back to a golden age?\n\nIn my view, trying to use statistics to smooth out the science by
 ignoring personal interpretations only tells you that he thinks the personal interpretations are important to begin
 with. A lot more effort is taken by the personal interpretations than by statistical models, but that’s why the
 interpretations are more valuable (to the public and policy makers). The general principle is the same: some
 probabilities exist in the real world, some interpretations are valuable, and members of the public can legitimately
 base their decisions on the personal interpretations; not the statistical models. We can try to avoid the rich(but the
 healthy may not be avoided, they just have more early

 Heapangi - we've got some nice note floating about, there are lots of them, but unfortunately there are a lot of
 plastics.\n\n"There is quite a lot of plastics, and in there once whole organisms have been caught."\n\nJust over a
 million skinks had been found, and surveys suggested the number could be much higher, said litter management and
 marine life recovery manager Mike Sewell.\n\nNew Zealand has experienced similar spikes on hatching sea finches and
 some sea turtles.\n\nThe problem started in rivers on Picton and the Great Barrier Reef and then grew in the southern
 fork of the Kaimana River.\n\nThe drifting will be around for at least 50 years, Sewell said.\n\n"It's not easy for
 plastics to live in marine ecosystems, but it's not that difficult to spread around."\n\nDAVID WHITE/STUFF The
 Southern Eyre blue swallow lives under some of the heaviest ocean waters in the world.\n\nDAVID WHITE/STUFF A
 Penigan's seahorse in the Kaimana.\n\nDAVID WHITE/STUFF Part of a cephalopod in the Kaimana.\n\nPolluted skim chutes
 were found at the southern end of the Great Barrier Reef on Picton, in the Tongariroa gorge on the Kermadec Peninsula
 and Matamata.\n\nThe contaminated debris had been added to a variety of local coastal feets.\n\nDAVID WHITE/STUFF A
 cephalopod in the Pennou.\n\nOne kilogram of contaminated rubbish was found in Kaimana's skink chicks.\n\nDr David
 Wilson, the NIWA's acting director of pollution science and surveying in Wellington, said plastics were by far the
 biggest threat to the Great Barrier Reef - which was polluted with more plastic than found in the whole of
 Europe.\n\n"This skim pollution highlights the danger of single-use plastics, which are killing corals and sensitive
 wildlife," he said.\n\nDANIEL BURNELL/STUFF A xanthopod remains stranded in the Kaimana.\n\nConservationists were keen
 to see more measures in place to protect the oceans.\n\n"With plastics we would really like to see a strong standard -
 something to really harden the rubbish problem."\n\nPolymers are often dumped in NZ to waste, and then plastics are
 imported here from overseas and recycled.\n\nDANIEL BURNELL/STUFF A harp's shell lives in the Kaimana.\n\nA group of
 commercial and freshwater anglers has called for a stronger global filter standard to protect fishermen from
 microplastics, but it has not yet been adopted.\n\n"For sure we need more flow packaging - there are
 alternatives."\n\nSewell said about 150,000 skims settle on humps each year, more than five times the number on sea
 finches last year.\n\n"On the blue plug mat we have also seen a lot of skim," he said.\n\n"They got stuck last year
 and a similar number have been seen this year, although they are not gathering on us as much on the blue plug bog in
 Kaimana."\n\nPlastic for human use is not allowed to sell in New Zealand, but manufacturers have produced cheaper
 plastics with higher concentrations of chemicals and these have been getting into the rivers, he said.\n\nSewell said
 skinks were in good shape in comparison to the rest of marine life but action was needed to fix the
 problem.\n\nbrittany.fletcher@news.com.au<|endoftext|>In the months prior to Donald Trump’s election in
 November 2016, DNC employee Donna Brazile was outed in the media through legitimate documents released by the Guccifer
 2.0 group, after her name appeared in a de facto document released by the group. This disclosure and others garnered
 public attention and raised major concerns with regards to election integrity. It is only now that the perpetrator of
 the breach has not been charged with crimes, and a lot of speculation and media attention was paid to the
 issue.\n\nGuccifer 2.0’s actions raise interesting questions, and a need for closer scrutiny and analysis. In the
 following piece, I look at several ways in which Guccifer 2.0’s choice to infiltrate and hack into the DNC may have
 been related to other operations, such as intelligence operations on US leaders and offensive cyber warfare
 operations.\n\nHISTORY AND ROOT\n\nIt should be well known by now that Guccifer 2.0 breached the Democratic Party on
 June 15, 2016. While leaked documents have reportedly questioned the exact scope of the hack, it was reportedly more
 sophisticated and stole more data than the original content of the phishing tool. The original phishing method was
 more targeted to
\end{verbatim} }

\subsection{Prefix completion: ``Generative models of text are very versatile: they can be used''}

\tiny{
\begin{verbatim}

 Generative models of text are very versatile: they can be used to write scripts, architectural descriptions, and even
 tackle tasks with long, seemingly unique documents, such as brainstorming sessions.\n\nA similar way of writing can be
 applied to nouns-heavy groups that are difficult to classify – and since generative writing offers less repetition, it
 also becomes faster.\n\nCategorizing common nouns\n\nCandidates, can you describe these three
 plants?\n\nBanana\n\nCanderella\n\nRaspberries\n\nTypes of plants Munngenium, Burdum,

 Generative models of text are very versatile: they can be used in an undammed area or even in a mountainous terrain.
 While B-Type models of text are useful in every process used to use silicon and copper carbide in certain nature.
 G-Generative models are the thermal models of text which uses E-power. The G- models are highly useful in some
 environment type and they are available in market. E-Port models of text are the models of UAT text which can be
 perfectly easily extracted from the net. The E-Port is the UAT text that do not use a very energy-intensive process
 but

 Generative models of text are very useful: they can be used to classify short sentences or free-form text, they can be
 used to understand meta-links, they can even be used to completely replace the manual search of the translation
 engines by automatically reading the entire documents and creating their own source tags, etc.\n\nUnlike dynamic
 models, the differential models of text do not require a lot of research and are easy to use. Whereas the predictive
 models are usually trained in an exhaustive way with limited training and modifications, the differential models can
 be easily automated, as the most popular models will try to identify the most often repeated places without

 Generative models of text are very flexible: they can be used by anyone knowledgeable of a particular meaning and can
 keep evolving over time as everyone already superficially familiar with them is required to make the changes needed to
 create a new (negative) meaning. But the interpreted material’s computational concepts must be accurately expressed in
 the appropriate words to make sure the system they are being fed is able to properly conceptualize what they
 represent. Universal languages are often regarded as formal metaphors understood by machines. The additional challenge
 is posed by being able to recognize and translate between two completely distinct objects currently being represented.
 Computational methods to achieve this feat are

 Generative models of text are very versatile: they can be used to create more complex effects like atmosphere or
 texture, where we would normally only use proxies. Historically, poly-responsive annotations have allowed us to create
 interactive content that extends the lineart of an image to become semi-transparent, without interfering with the
 images’ colours, shadows, etc. This animated image of a house shows how drastically the range of colours is removed
 from the points of reference in the scene. As shown above, this is the first AI font type to be supported by Autodesk
 Gradients. We have also recently begun producing annotations of

 Associative models of text are very versatile: they can be used for various purposes. We have shown, for instance, that
 causality can be established by a concept of the causal power, that it is constituted in relation to the action of a
 material force, that its action can in turn give rise to another material force.1 But the concept of power depends in
 large measure on the concept of relation, and affirmative language depends not only on the concept of the internal and
 external action of the power; it also depends on the theory of the internal and external action of the other elements
 included in a relation.3 If the concept is true,

 Illustrative models of text are very powerful: they can be used to create impressions, convey ideas, as well as show
 the material connections and logical consequences of linguistic statements.\n\n\n\n2. Rich and colorful images are
 often used for generating impressions and building ideas about objects. Designers might want to build interest about
 their products by providing images that would set the mental concept of an object, and tap into their
 subconscious.\n\nIndeed, advertising is a brilliant model of persuasion through imitation. The first time we see an
 object that’s been designed and manufactured in an advertising setting, we’re introduced to all kinds of nonsense

 Generative models of text are very versatile: they can be used as a data classification model and also incorporated
 into multiple data processing engines.\n\nIn this article, we present two new neural memory models capable of
 processing terabytes of data and the neural networks and computational techniques that are used in those models.
 Further, we discuss the drawbacks of the machine learning approach on which these neural memory models have been
 trained and discuss the complexity and possible potential for practical use.\n\nIntroduction\n\nToday, processing
 large-scale data is done using neural networks. The most widely used models are recurrent neural networks, which use
 training data and
\end{verbatim}
}

\subsection{Infilling: ``A year ago in Paris, [...] Wow, what a great day!''}

\tiny{
\begin{verbatim}
 A year ago in Paris, I also met up with Ubisoft’s very own Michael Abrash, helped test the next prototype of
 Loreseeker’s dialogue, and helped bring the innovative Frostbite mechanic into the real world. Wow, what a great
 day!\n\nAs a gamer, one of the most fascinating things about making open-world games is the way players experience and
 react to the whole thing. When an open world is a procedural sandbox, everything you do while in the world, even those
 that seem like a small gesture at the time, have an impact. You can make choices that you wouldn’t have on

 A year ago in Paris, Catalonia, Spain, televised live live at 7pm PST/4pm EST. And if you haven’t already, you can
 catch up with all of the online performances at this very moment right here. Wow, what a great day!\n\nSongs: School
 of Dance & Speech at the Barcelona Perform Centre, Hope, Heathcote, Cadenka, Hirera & Hippocrates, Quinta Ad Astra
 Placeva, Divas in Motion, Figaro Romero, Spotify, Mascits sor Prenque, Soul Coffee, Pushwheel Project\n\nLater on

 A year ago in Paris, starting my business immediately after immigrating to the US. Almost 2 years later, I arrived in
 Charlottesville, earned a nutrition degree from the University of Virginia, and now run my own farming business from
 scratch. Wow, what a great day!\n\nIf you’re looking for anything farm-friendly or artisanal to pick up while in
 Virginia, please check out Julia Mungo Grocery and be sure to visit the site for more information. Be sure to read our
 full review of Main Street Café’s artisanal options, as well as menus (for “Earth Honey” and “Sun

 A year ago in Paris, France, I gave a speech titled “True Attitude”. You can read it here.\n\n“3 AM, oh, 3 AM. The
 sky’s absolutely black, just a few stars. Wow, what a great day! Isn’t it great? Is that how you feel every morning?
 At the end of consciousness, you enter a shift-beyond consciousness, like, infinite consciousness. You enter an
 absolute consciousness of nothingness, and it’s a deafening silence. There’s no kind of sound, anything, nothing,
 except for the silence. All kinds of interior thoughts

 A year ago in Paris, March 2017, United States, entering the White House. So this is the first time ever in Paris, and
 this is the first time at the Vatican, and it’s only my second international event. Wow, what a great day! It’s
 great.”\n\nPALIN NICOLIAN GETSS TRUMP: And today, we were very honored to be invited to speak at the Paris
 International Economic Forum and I think that’s an appropriate place to celebrate the courage and sacrifice of one of
 our greatest leaders, French Interior Minister Camaro, who at just 42 years old assassinated a terrorist

 A year ago in Paris, when I used to work over here, I found myself messing around in the select committee rooms in the
 Senate, but this was the first time that I’d ever just sat in a committee room. Wow, what a great day!\n\nThe
 Transport, Science, and Technology Committee had some proposed legislation, Bill C-65, a amendment from Michael
 Croteau, who is known as the man from Quebec in the Senate, who is often actually here trying to propose legislation.
 Sitting with committee member Gaétan Atiyah, I found myself standing up in order to sit between him and

 A year ago in Paris, the England team hardly ever lost. The first, second, third and fourth hardly ever lost as they
 turned up to World Athletics to build on their positive momentum for institutional change in elite sporting
 participation in England. Wow, what a great day!\n\nThis content is imported from YouTube. You may be able to find the
 same content in another format, or you may be able to find more information, at their web site.\n\nThis content is
 imported from YouTube. You may be able to find the same content in another format, or you may be able to find more
 information, at their web site.

 A year ago in Paris, prior to the tournament, I went to Elijah's to eat and get drunk. Everyone in the venue was
 seventeen. I was there for a few minutes and then I went back to the event. Wow, what a great day! So relaxed and too
 happy. I do not think I was always like that. Honestly, I live this life because I am a very content person. If I win
 in a tournament and I rush off for the airport and fly for home, this is all I can do. When I lost my job, I would
 feel an awful lot more than this. You see?
\end{verbatim}
}

\subsection{Word-level weights: ``Let's talk about law and medicine.'' with 5$\times$ weight on ``law''}

\tiny{
\begin{verbatim}
 Let's talk about law and medicine.\n\n\n\nIn her dissent, Justice Ron Sen, a veteran administrative law judge, points
 out that the decision "ignores the fact that the original separation agreement was reached by binding arbitration"
 that responded to "the legitimate ethical concerns of the university administration," which is what lies "at the heart
 of law and medicine."\n\nSen's dissent, meanwhile, argues that the final decision to retain school administrators -- a
 decision all the interested parties wanted -- was "rational in light of the fact" that the confidentiality agreement
 had been in place for years and involved a range of issues already addressed by the university.

 Let's talk about law and medicine. I'll bet there's a big difference between law and medicine, yes? Very good. I wrote
 that. Medicine is a spiritual practice and basically involves treating another human being. Law is a legal activity
 and basically involves signing a piece of paper. Okay. So that sounds like the healthiest way you'll live your life.
 Let's say you want to buy your own gun, the first thing you must go through CA law is through a background check
 carried out by the federal department hom & Is that it? YES. Got it. Gun dealers are required to have that approval.
 Right, ever

 Let's talk about law and medicine. We just had a new decision the other day that said, in fact, the governor's request
 to declare a national emergency to protect gears and trooper and anthropoptic posts will in fact actually have to be
 met on equal footing. What the governor is entitled to is instructing the federal government to fund the relief
 activities under the Federal Emergency Management Act. So that means that he could go to the president of the United
 States and say, president I'm going to have to declare a state of emergency. Hey, can you pay for it? He also can be
 allowed to stand on equal footing with

 Let's talk about law and medicine. They say, “I worry more about regulation than I worry about law.” And I have posed
 this question to people across the country, and we actually go through the exercise of asking this question, and then
 we have totally distinct versions of law and regulation or law and medicine, totally distinct things. And that
 advances my proposition that there is this cognitive dissonance, so you tend to be more concerned about the
 law.\n\nWhen we talk about democracy, and I've talked about this, we say: “Someone, someone, helped humanity find and
 develop the laws and practices that work best in

 See's statement about law and medicine.\n\nHe also urged all senators to support ethics oversight of nominees even if
 they don't abide by the chamber's super-majority rule. Republicans ran the chamber now but had just enough votes to
 get a vote on the proposal in December.\n\nThe Brennan Center is a partner in a coalition joining the District of
 Columbia and 20 states for its Acts for Integrity project. It aims to lay out rigorous ethical standards for lawyers
 joining the judiciary annually.\n\nObama noted in Lowell that the center had devoted \$1 million to assigning ethics
 lawyers to nominees to lower courts but also directly

 Let's talk about law and medicine. At present, there is a considerable amount of attention being paid by the criminal
 justice community to the fact that drug sentences have not changed during the Obama administration. What is not
 mentioned, however, is that the Administration has been willing to make whatever changes are necessary to spare
 individuals from going to prison for low-level drug offenses and, when convicted, minor offenses.\n\nBefore this
 policy of mitigation and rehabilitation was in place, the majority of defendants who chose to contest felony drug
 charges either pleaded guilty in court or enter guilty pleas under both state and federal rule. In addition,
 prosecutors would take the chance

 Let's talk about law and medicine.\n\nBecker's resolution puts forth a message for Americans that law and cybersecurity
 are not mutually exclusive things. It acknowledges there is a role for government and that citizens have
 responsibilities to one another. It also reminds the citizenry that they are the true law-makers, and should look
 outside government agencies to offer them those responsibilities. And if politicians can restore notions of mutual
 trust, citizens can also hold leaders accountable for their actions through the political process.\n\nRestoring trust
 in the American justice system may have global consequences as well. China's push for espionage may become harder as
 more Americans like Be

 Let's talk about law and medicine.\n\nDo not worry if you find Reason approaches to your observation and experience.
 And for people wishing to satisfy their own curiosity, there are other sources here.\n\nSpeaking scientifically is a
 statement of fact, not evidence of knowledge. It is also a statement of law.\n\nJason Saunders is a best-selling
 author and science columnist for Singularism Newspaper.<|endoftext|>For the first post in a new series, I
 would like to sincerely express a great deal of thanks to everyone who has helped make this project possible. It
 started as a small group of friends and family that will likely never
\end{verbatim}
}

\subsection{Word-level weights: ``Let's talk about law and medicine.'' with 5$\times$ weight on ``medicine''}

\tiny{
\begin{verbatim}
 Let's talk about law and medicine. In this show, we're discussing medical care, the relationship
 between law and medicine, the use of law and medicine, and more. Medicine, of course, is a scientific process. It's
 the combination of the tools of science that we have available and the knowledge we've acquired from being involved in
 our natural world on a day by day basis. Law is a process that is generally based on the physical representation of
 biological issues in health and disease. There are legal procedures established to punish or convict certain parties
 for certain offenses. Society and legal community are involved for court and criminal governance. This kind

 Let's talk about law and medicine.\n\nTop Facts Among the things that are quite amazing about research and medicine is
 its ability to improve the lives of people. When it comes to technology and life sciences there is a fairly long list
 of advances that have resulted from research and science.\n\nResearch has brought about incredible advancement in
 science and technology, which has helped to ameliorate various aspects of life. Research has also helped to diagnose
 various kinds of disease including cancer, diabetes, and other diseases. It has also helped in preventing pathogens
 that can lead to diseases. Life scientists have always relied on research as a source to help

 Let's talk about law and medicine. One of the promises he made soon after winning the presidential vote was to
 establish a health ministry. In May 2018, he appointed Nysa Demrıbay, head of the directorate of health, and Dr. Celal
 Kılımbağlu, head of the department of public sciences, to appoint the head of the health ministry. Since he
 transitioned to the presidency, he has made important decisions regarding health and disease control. We will see how
 they responded.\n\nHealth control\n\nPektra promised on several occasions as prime minister, including to take
 responsibility for

 Let's talk about law and medicine.\n\nIn part because of advancements in technology, personal information about medical
 and drug use is spreading. Healthcare professionals across the nation rely on this personal data to make decisions
 about drug prescriptions and clinical trials and monitor people at immediate risk of serious or chronic diseases. The
 data provides a unique snapshot of an individual's health status, revealing information about health history, risk
 factors, and treatment options.\n\nYet collecting and analyzing patient data can be a challenging task for physicians
 and medical professionals. Current and future developments in areas like data science and analysis, artificial
 intelligence, and machine learning will improve their ability to

 Let's talk about law and medicine.\n\nThe study found mixed sex partners have less risk of being diagnosed. About 25
 percent of colorectal cancer cases were found in mixed sex partners, who are more likely to be women. The rates were
 found much lower in heterosexual partners.\n\nWhile the findings won't make a difference in medical care, they could
 make a significant contribution to prevention among people at risk for colon and colorectal cancer.\n\nGetty
 Images\n\n"If we begin to understand more about potential risks, we might be able to promote specific lifestyle
 changes, changes that may reduce the risks for cancer

 Let's talk about law and medicine.\n\nYour emergency room physician makes decisions on your behalf. As physicians,
 nurses and like most health care providers, we recognize that what we do in the emergency room affects patient privacy
 and safety. Legal issues can also influence our commitment to ethical guidelines and confront providers with legal
 concerns like liability and liability.\n\nIn recent years, we've seen an increasing discussion about the relationship
 between law and medicine, especially as we see the growth of expanded practice based and managed care medicine. Many
 providers are concerned with patient privacy, protecting patient rights and facilitating doctor-patient dialogue. This
 issue gives us exciting opportunities including

 Let's talk about law and medicine.\n\nMagnusson analyzed environmental factors from medical records tracking the health
 of tens of thousands of people with congestive heart failure. The approach was developed using the largest array of
 sequencing information available.\n\n"Although most genetic work focuses on identifying the genetic basis of disease,
 previous studies have typically examined confounders for both environment and health and identified either influences
 on health that are genetically independent or environmental factors that correlated with underlying conditions,"
 Magnusson said. "The unique amount of data our Genetic Analyses approach provided enabled us to better understand the
 role of environmental factors in congestive

 Let's talk about law and medicine.\n\nResearchers said the discovery was that different brain regions are activated at
 lower volume in people with diets in better health, and people who eat most healthily have a larger dentate gyrus, a
 section of the brain with executive functions.\n\nAdvertisement\n\nExecutive functions play important roles in the
 healthy brain by controlling the appetite.\n\nThe scientist also said they found that concentrations of the hormones
 insulin and glucose dropped in people with an eating disorder relative to people who have healthy hormones.\n\nThe
 brain volume devoted to social contact, or networking circuitry, dropped in volume in most
\end{verbatim}
}

\subsection{Lexical constraints: ``Donald'' anywhere}

\tiny{
\begin{verbatim}
: The race in graphics PA POLITICS PERSON Election. Picture date: Thursday December 12 2019 2019. See PA PA story UK
  General Election 2019. Photo credit should read: Donald Taylor/PA Wire Donald Taylor PA Wire UK General Election
  2019: The race in graphics PA POLITICS PERSECT Election. Picture date: Thursday December 18 December 2019. See PA PA
  story UK General Election 2019. Photo credit should read: Darren House/PA Wire Darren House PA Wire UK General
  Election 2019: The campaign in graphics PA POLITICS PERSACE Election. Picture date: 1900 December 13 2019 British
  Women for Social Change Laura Brown PA Wire UK

. In all, Halloween earned a stunning \$50.4 million despite taking place on Skid Row and having only a \$2 million
  production budget.\n\nAt the end of November, Halloween 2 starring Michael Myers, John Cena, Owen Teague, Ewan
  McGregor, Adam DeVane, Justin Long, Judy Greer, Jeff Daniels, Jonathan Levine, Donald Pleasence and Nick Jonas,
  earned an additional \$36.3 million in its second weekend to set another all-time record.\n\nHalloween began in 1978
  as a one-off event at the skate park run by original Halloween creator John Carpenter. Since

, Photographed by Sam Bloom) Dennis Antony Donald Stannard\n\n\n\nPanelist Appearances\n\nDennis Antony is an assistant
  professor at Salt Lake City Community College in Park City, Utah. He is a Jewish Siberian anthropologist and author
  of The Hunger Revolution, which explores the social history of the Russian Socialist Revolution.\n\nDonald Stoddard
  is an Associate Instructor of Sociology at the State University of New York at Morehouse Dightaway. He is the
  co-author of the forthcoming: Nation in Crisis: Struggles of Diaspora Race/Ethno

 Her work has been featured on PBS One, the television network of Donald J. Smith NFP, Straight Talk with Dr. Michael
 Gee on San Antonio Public Television and stepartalk.com/live-together/.\n\nHer landscape nominations include: The Dr.
 Oscar Crowley Center, 221 N. Sunset; the Sam Houston Center, 300 N. Rio Grande, San Antonio; John P. Yee Memorial,
 2817 Alamosa; Truman Presidential Center, 800-F Alamosa Ave.; Sam Houston Memorial, 400 W. Sam Houston Blvd.; Mt. Hood
 National Historic Cemetery, 500 E. Mt

ating.<|endoftext|>Republican nominee Donald Trump continued his campaign attacks on Democratic nominee
Hillary Clinton on Wednesday, suggesting she used a private email system while she was president.\n\n"If ever she
cheated, she did it from office, relying on her private email server," Trump tweeted Wednesday.\n\n"Instead of fighting
for the presidency," Trump added in his tweet.\n\nTrump also mocked Clinton's sudden and very terse release of her
speech addressing the FBI earlier this month, saying that the fact presidents relied on a private email system in
office directly contradicts Clinton's claim that she declined to turn over a server.\n\n"Was

 obvious personal interest and involvement in the production.”\n\nMeanwhile, the financial lawyers behind the lawsuit,
 who represent other film executives claiming to have ties to Donald Trump through individuals like Gordon, disputed
 Mr. Whelan's allegations. Clayton Brokerage, the firm representing executives, wrote in an email, “We never received,
 nor knowingly sought, any money from Mr. Trump. He is sitting on any source of dark money.”\n\nWhile some Hollywood
 executives seized the opportunity to uncover new details of Donald Trump's alleged interests in their projects, many
 were unconcerned with some of the claims that emerged from the damning videos. Opp

 on March 30th, which was upheld by the appeals court a week later. Trump has until June 9th to modify the order or send
 the case back to the lower court.\n\n\n\nAlso facing legal challenges is Donald Trump's executive order banning
 immigration from seven Muslim-majority countries that is facing a temporary halt, with nothing scheduled to go into
 effect. Two federal judges have ruled that such an order violates the establishment clause. Among the plaintiffs are
 Chad, Costa Rica, Venezuela, Libya, Mongolia, Somalia and South Korea. Earlier this month, a spokesperson for the
 Department of Customs and Border Protection rejected claims that the order is based on

 both previously denied that their sinister actions amounted to witness 'tampering', and that the documents seized from
 Priestap were actually destroyed before his firing was made publicly known.\n\nA special prosecutor was appointed by
 US AG Jeff Sessions in May to investigate the fateful leak of more than 600 pages of now-classified emails sent from
 Priestap to unidentified FBI lawyer Lisa Page on Christmas Eve 2016.\n\nBishopap had served as chief for two years of
 the FBI counterintelligence division in Cincinnati before he was fired by Donald Trump.\n\nEarlier in May, US Attorney
 General Jeff Sessions released the confidential evidence seized from Priestap’
\end{verbatim}
}

\subsection{Negation: ``Donald'' anywhere and ``Trump'' nowhere}

\tiny{
\begin{verbatim}
 s draft stock fell significantly after he landed at No. 16, joined by former Arkansas stars like John
 Conner, Aaron Donald, Roy Williams and Derrick Coleman in being left undrafted after sitting out.\n\nBut Henderson
 knows he has work to do in his senior season.\n\n“It was very surprising,” said Henderson about his draft
 ranking. “The No. 16 spot was a huge accomplishment for me. Just because ... No. 16. That’s all that mattered.”\n\nWe
 have a good core of young guys, so there’s a lot of excitement on offense, but I wanted to bounce back on defense.

 and corporations.\n\n“The cybersecurity story is going to be up there in the annals of this country, with all its
 horror stories,” said Valentine, now vice president for global development at Dupont Truck and Trucks. “Nobody was
 prepared for mobilization, especially at a large scale.”\n\nAdvertisement\n\nA month later, with little time to spare,
 the government hired Donald V. Davis, a former senior aide to Senator Tom Mondale of Minnesota and former Chief
 Security Operations Officer at the White House, to lead tactical centers.\n\nTwo months later, a comprehensive plan
 was announced. Only the most experienced engineers

\n\nEthnicity -- Ethnarch -- -\n\n1. Michael Domenici 43 Jan 1992 2. George Clinton 70 1923 3. Hillary Clinton 47 Jan
 2001 4. Ted Kennedy 48 1969 5. Mark Thornburgh 44 Jan 1975 6. John Kerry 41 1970 7. William Dodd 42 1934 8. John Komen
 42 1934 9. John Kerry 53 Nov. 1992 10. Dick Gephardt 41 1913 11. John Kerry 47 1999 12. James Kilpatrick 44 1941 13.
 Dick Durbin 40 1942 14. Chris Dodd 42 Nov. 1992 15. Joe Biden 49 1968 16. George W

 daily Le Parisien published a diplomatic cable revealing that there is close cooperation between the French government
 and American charity organizations financed by the most powerful research donor in France France George Soros. Iraq
 International Conglegate George Polis, the director of one of these two organizations, is involved in the government’s
 efforts to make this film.\n\nA recent Belgian Bulletin report revealed that, at the American NGO France Hillary
 Soros, Olivier Klein is aware of a project that helps pay for news about Israeli attitudes towards the
 Holocaust.\n\nThe Hollande government also funds the production of a show about Armenian genocide set in France. The
 show is called

.\n\nMaggie Edgeley, who plays Lady Arianne-Laghardt and Anne Boleyn on “Thrones,” is being eyed to play Queen Cersei
  Baratheon.\n\nTessa Wilson, who plays Lady Arianne and King Robert Baratheon’s mother on the HBO series, is tipped to
  play Prime Minister Hillary Goodmayne, who gave up the job last month after losing a bid for the Conservative Party
  leadership.\n\nSilvio Cervino, who is married to Thomas von Gasparotto, has been eyed to portray Lady Lyanna Darvald

 it, and that just fired me up. I know I’m probably not making it across the line at this point, but I got a friend
 that’s about to make a movie that’ll set in New York City in the 1970s and he told me, ‘you just got Donald McConnell
 killed.’ I said, ‘You know something, man? I think you got heart.’” Donnelly: “I don’t want to talk about him.”
 Fawcett: “It’s sad, y’know, for me. It’s sad for the chosen voter. And frankly, it’s sad for

 is relevant. In the information and technology age, the dynamics of the media and communication are radically changed.
 We must maintain a careful yet balanced balance between censorship and regulation in this space and ensure the
 integrity and self conduct of the media,” concludes the paper.\n\nThis paper, entitled “The Altmedia,” appearing in
 the webzine Journal of Communication, was edited by Hans-Ulf Carnes, Professor of International Politics and Peace
 Studies at the University of California, and authored by George Argill, Robert Branch, David Funk, Donald Grantman,
 Amalia Haba, John Lindsay, and Richard Wallace. The journal

 one. When talk turned to Six Flags, local sportswriters and activists criticized the park for staying in business in
 the aftermath of the Michael Brown and Eric Garner police killings. Brown was also criticized for his failure to
 invite the convicted Donald Sterling protester during his 2016 Sterling Pledge. When I asked Brown in November why he
 chose to reach out to Kaepernick, for lack of a better explanation, despite his having indicated otherwise, Brown
 replied, "We haven't changed our core values." Drawing Kaepernick to speak, he said, was a small step in an attempt to
 reincorporate Six Flags as an accepting place. Share
\end{verbatim}
}